\documentclass[manuscript,screen]{acmart}

\AtBeginDocument{%
  \providecommand\BibTeX{{%
    \normalfont B\kern-0.5em{\scshape i\kern-0.25em b}\kern-0.8em\TeX}}}

\setcopyright{acmlicensed}
\copyrightyear{2018}
\acmYear{2018}
\acmDOI{XXXXXXX.XXXXXXX}

\usepackage{algorithmic}
\usepackage{graphicx}
\usepackage{textcomp}
\usepackage{xcolor}
\usepackage{booktabs}

\usepackage[most]{tcolorbox}
\usepackage{mathrsfs}
\usepackage[ruled,linesnumbered]{algorithm2e}
\usepackage{multirow}
\usepackage{pifont}
\newcommand{\set}[1]{\ensuremath{\mathcal{#1}}}

\setcopyright{none}
\renewcommand\footnotetextcopyrightpermission[1]{}
\usepackage{subcaption}

\usepackage{eso-pic}
\usepackage{lipsum} 

\AddToShipoutPictureBG*{%
  \AtPageUpperLeft{%
    \setlength\unitlength{1in}%
    \raisebox{-2cm}[0pt][0pt]{
      \makebox[0pt][l]{\hspace*{2.6cm}\parbox{\textwidth}{ \fontsize{8}{11}\selectfont Accepted to the ACM Transactions on Privacy and Security on 2 December 2024\vspace{-0.3cm}\\\rule{\textwidth}{0.4pt}}}%
    }%
  }%
}

\begin{document}

\title[Leveraging Domain Constraints for Robust Malware Detection]{Level Up with ML Vulnerability Identification: Leveraging Domain Constraints in Feature Space for Robust Android Malware Detection}

\author{Hamid Bostani}
\email{hamid.bostani@ru.nl}
\affiliation{%
  \institution{Digital Security Group, Institute for Computing and Information Sciences, Radboud University}
  \city{Nijmegen}  
  \country{The Netherlands}
}

\author{Zhengyu Zhao}
\email{zhengyu.zhao@xjtu.edu.cn}
\affiliation{%
  \institution{Faculty of Electronic and Information Engineering, Xi'an Jiaotong University} 
  \city{Xi'an}
  \country{China}}
\email{zhengyu.zhao@xjtu.edu.cn}

\author{Zhuoran Liu}
\email{z.liu@cs.ru.nl}
\affiliation{%
  \institution{Digital Security Group, Institute for Computing and Information Sciences, Radboud University}
  \city{Nijmegen}  
  \country{The Netherlands}
}

\author{Veelasha Moonsamy}
\email{email@veelasha.org}
\affiliation{%
 \institution{Horst Görtz Institute for IT Security, Ruhr University Bochum} \city{Bochum} 
 \country{Germany}}

\renewcommand{\shortauthors}{Bostani, et al.}

\begin{abstract}
Machine Learning (ML) promises to enhance the efficacy of Android Malware Detection (AMD); however, ML models are vulnerable to realistic evasion attacks---crafting 
realizable Adversarial Examples (AEs) that satisfy Android malware domain constraints. To eliminate ML vulnerabilities, defenders aim to identify susceptible regions in the feature space where ML models are prone to deception. The primary approach to identifying vulnerable regions involves investigating realizable AEs, but generating these feasible apps poses a challenge. For instance, previous work has relied on generating either feature-space norm-bounded AEs or problem-space realizable AEs in adversarial hardening. The former is efficient but lacks full coverage of vulnerable regions while the latter can uncover these regions by satisfying domain constraints but is known to be time-consuming. To address these limitations, we propose an approach to facilitate the identification of vulnerable regions. Specifically, we introduce a new interpretation of Android domain constraints in the feature space, followed by a novel technique that learns them. Our empirical evaluations across various evasion attacks indicate effective detection of AEs using learned domain constraints, with an average of 89.6\%. Furthermore, extensive experiments on different Android malware detectors demonstrate that utilizing our learned domain constraints in Adversarial Training (AT) outperforms other AT-based defenses that rely on norm-bounded AEs or state-of-the-art non-uniform perturbations. Finally, we show that retraining a malware detector with a wide variety of feature-space realizable AEs results in a 77.9\% robustness improvement against realizable AEs generated by unknown problem-space transformations, with up to 70$\times$ faster training than using problem-space realizable AEs.
\end{abstract}



\maketitle
\section{Introduction}
\label{section:introduction}
Due to the ongoing proliferation of Android malware, the application of Machine Learning (ML) for Android Malware Detection (AMD) continues to remain a topic of interest for security researchers~\cite{b6,b7,b8,b9,b10,b11,b12,b13,b14}. However, ML-based solutions are vulnerable to evasion attacks that generate adversarial examples (AEs)~\cite{b22}---malware that is crafted to purposely be misclassified as benign. These adversarial attacks exploit blind spots within feature space (i.e., the model's decision space) where the decision boundary is not accurate due to insufficient training samples~\cite{b15}. To confront evasion attacks, various defense strategies have been proposed to either detect (e.g.,~\cite{carrara2018adversarial,shan2021using,grosse2017statistical}) or eliminate (e.g.,~\cite{b33,b41,incer2018adversarially}) blind spots; nevertheless, in the Android malware domain, defenders must cope with realistic attacks that capitalize on \textit{feasible} blind spots. Indeed, the entire blind spots do not show vulnerable regions in the context of malware because realistic evasion attacks solely target blind spots within feasible regions where feature representations of feasible apps can settle. The primary solution for uncovering vulnerable regions entails exploring realizable AEs, such as leveraging them in \textit{adversarial hardening}~\cite{b42}, encompassing methods that integrate AEs into the training process. However, generating realizable AEs is challenging as AEs must be \textit{feasible apps in the real world}, meaning they must satisfy the \textit{domain constraints}~\cite{b39,b40} (e.g., preserving malicious functionality~\cite{b17}).

\begin{figure}[!t]
    \centering
    \includegraphics[width=0.25\columnwidth]{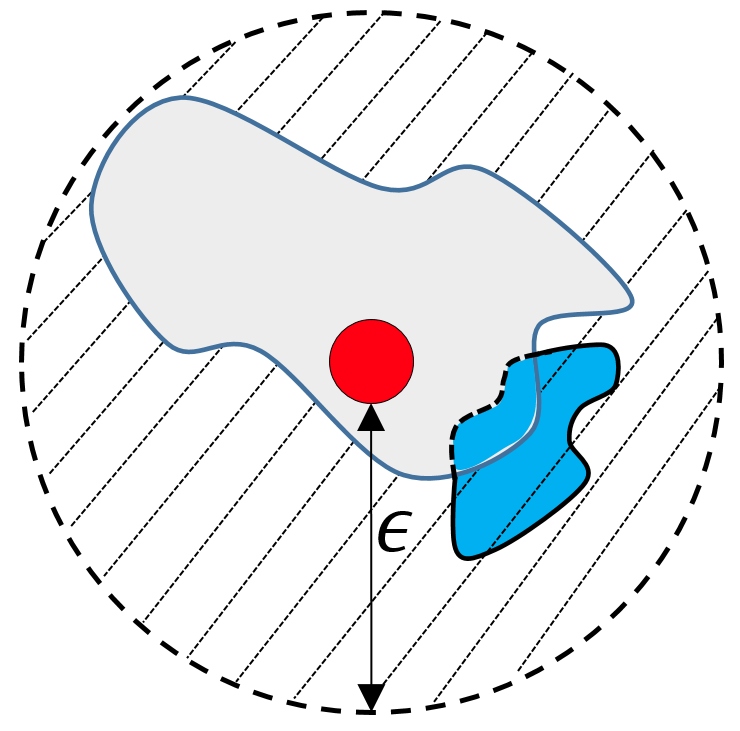}
        \caption{Feature space achieved by existing unrealistic attacks (blue) may not cover the realizable AE space (gray). The $\epsilon$-ball covers all possible AEs that can be generated for the malware sample (red).}
    \label{fig:decision_boundaries}
\end{figure}

Generally, there are two different approaches to generating AEs. The first approach is to generate norm-bounded AEs by modifying specific features that can best mislead the detector without considering domain constraints~\cite{b19,b28,b43,b44,b45,b46}. 
This approach is moderately efficient and can offer some degree of adversarial robustness when utilized in adversarial hardening because the feature space of realizable AEs is just a sub-space of the full space of AEs~\cite{doan2023feature}. However, AEs generated by this approach in the Android domain~\cite{b1,b87}, might not be realizable AEs, as the resulting AE space fails to fully cover feasible regions that are vulnerable to realistic evasion attacks~\cite{b42}, as illustrated by Figure~\ref{fig:decision_boundaries}. For instance, the attacks proposed in~\cite{b21,b29} might not always generate realizable AEs as the adversarial features added to the Manifest files of apps can be removed by pre-processing operators~\cite{b17}. Figure~\ref{fig:decision_boundaries} also indicates that exploring the full space of norm-bounded AE space is not necessary if we can model the smaller space of realizable AEs.

The second approach is to directly generate realizable AEs in the problem space, i.e., applying problem-space transformations into malicious apps that induce realizable perturbations in the feature space~\cite{b1,b17,b30,b48,b49,b50}. Although this approach fundamentally ensures that the domain constraints are satisfied, we find it to be sub-optimal for adversarial hardening mainly due to three reasons. First, applying problem-space transformations is computationally expensive~\cite{lucas2021malware} whereas perturbing feature vectors is typically simpler and more efficient than manipulating objects in the problem space~\cite{b39}. Second, finding effective problem-space transformations is challenging because they must not only mislead the detector but also meet domain constraints. 
For instance, the transformations utilized in~\cite{b29,b30,b81} fail to meet the domain constraints because they either can be thwarted by removing newly-added content through pre-processing~\cite{b29,b81} or result in functional disruptions~\cite{b30}.
Third, the problem-space transformations used in adversarial hardening might be ineffective to unknown attacks, implying that attackers utilizing new transformations could still successfully evade detection~\cite{lucas2021malware}.

To tackle the challenges associated with identifying vulnerable regions, this paper aims to help defenders uncover such regions by exploring the properties of feasible apps within the feature space. These properties, which represent domain constraints in the feature space, assist defenders in pinpointing feasible regions that might be susceptible to realistic evasion attacks. For instance, by leveraging feature-space domain constraints, adversarial hardening can harness the advantages of both the feature space and the problem space when generating AEs, i.e., being efficient by directly modifying features while satisfying the domain constraints. To this end, we first interpret the domain constraints of Android malware in the feature space (\S\ref{section:problem-space_constraints_in-the-seature-space}), then learn domain constraints from the feature representations of a large number of apps (\S\ref{section:learning-problem-space-constraints}), and finally apply them to counter evasion attacks (\S\ref{section:applying-the-problem-space-constraints}). More concretely, we first argue that Android domain constraints are meaningful feature dependencies that exist within the feature space. This implies that feature-space AEs are realizable when they adhere to these feature dependencies. Then, we introduce two sets of dependencies over the feature values, named \textit{perfect} and \textit{relatively strong} feature dependencies, which can represent domain constraints in the feature space.
Next, we present a domain-constraint learning method to extract meaningful feature dependencies. Specifically, the proposed method utilizes statistical dependencies and Optimum-path Forest (OPF)~\cite{b55} to learn domain constraints from the feature representations of training samples. Here, OPF, which is a graph-based pattern recognition method, is adapted to extract meaningful feature dependencies. Finally, we apply our learned domain constraints across various defense methods to illustrate their effectiveness. In particular, we propose an AE detection method to preemptively identify AEs by differentiating them from feasible apps using our learned domain constraints. Our empirical evaluation shows that our proposed method can successfully identify 89.6\% of AEs generated by various evasion attacks. Moreover, we incorporate feature-space realizable AEs into Adversarial Training (AT)~\cite{b33} to enhance the robustness of Android malware detection against realistic evasion attacks, generating problem-space realizable AEs. Such feature-space realizable AEs are generated during AT by considering not only the norm-bounded constraints but also our learned domain constraints. Our empirical analyses on DREBIN~\cite{b9}, DroidAPIMiner~\cite{b11}, RAMDA~\cite{li2021robust}, and R-PackDroid~\cite{scalas2019effectiveness}, four different malware detectors, reveal that our defense outperforms both AT based on norm-bounded AEs (9.3\% over DREBIN, 34.5\% over DroidAPIMiner, 20.0\% over RAMDA, and 8.1\% over R-PackDroid) and state-of-the-art AT based on non-uniform perturbations~\cite{b38} (4.7\% over DREBIN, 16.6\% over DroidAPIMiner, 11.1\% over RAMDA, and 3.1\% over R-PackDroid).
Our evaluation also highlights the better performance of our defense than problem-space realizable AEs in efficiency and generalizability for adversarial retraining~\cite{b54}.
Our contributions\footnote{We make our code publicly available at {\url{https://github.com/HamidBostani2021/robust-Android-malware-detector}}.} can be summarized as follows:
\begin{itemize}
\item We propose a novel interpretation of domain constraints in the feature space for AMD (\S\ref{subsec:problem-space_constraints_interpretation}). This new interpretation considers key feature dependencies of feasible Android apps in feature space to specify feasible regions where the feature representations of realizable AEs may reside. We then propose a novel domain-constraints learning technique based on the statistical correlations between features and a graph-based clustering algorithm called OPF to extract meaningful feature dependencies from large-scale data (\S\ref{sec:our-learning-method}).

\item 
We demonstrate how these learned domain constraints can be utilized either to identify AEs, which are not feasible apps (\S\ref{subsection:ae_detection}), or to generate feature-space realizable AEs for adversarial hardening to improve the robustness of AMD against realistic evasion attacks (\S\ref{subsection:adversarial_hardening}).
\item 
We empirically evaluate the proposed AE detection with three evasion attacks, GenDroid~\cite{xu2023gendroid}, ShadowDroid~\cite{zhang2021shadowdroid}, and Grosse Attack~\cite{b29}, demonstrating that our defense can successfully identify their generated AEs. Furthermore, our extensive experiments on four different Android malware detectors, DREBIN~\cite{b9}, DroidAPIMiner~\cite{b11}, RAMDA, and R-PackDroid~\cite{scalas2019effectiveness} demonstrate that our defense provides superior model robustness than AT based on norm-bounded AEs and the state-of-the-art defense based on non-uniform perturbations~\cite{b38}.

\item We validate both the efficiency and generalizability of our defense over adversarial retraining based on problem-space realizable AEs~(\S\ref{sec:feature_vs_problem_space_comparison}).
\end{itemize}
\section{Related Work}
\label{section:related-work}
In this section, we first provide an overview of prior studies that have explored AEs in the Android domain, either within the feature space (\S\ref{subsec:feature_space_AEs}) or problem space (\S\ref{subsec:problem_space_AEs}). Furthermore, we review recent studies that explore realizable AEs with feature-space domain constraints but in domains other than Android (\S\ref{subsec:feature_space_RealAEs}).

\subsection{Feature-Space AEs}
\label{subsec:feature_space_AEs}
There exist a large body of related work~\cite{b20,b21,b29,b44,b43,b45,b46,b62,b72,xu2023gendroid} that investigated feature-space AEs. Xu et al.~\cite{xu2023gendroid}  introduced a black-box attack, incorporating the attention mechanism and the Jacobian-based saliency map algorithm.
Croce et al.~\cite{b62} proposed a query-based evasion attack using random search and evaluated it in various contexts, including AMD. Xu et al.~\cite{b46} developed a semi-black-box framework based on the simulated annealing method to perturb features of Android apps by querying the target malware detector. Li et al.~\cite{b21,b20} proposed gradient-based and gradient-free evasion attacks to generate AEs in the feature space. Rathore et al.~\cite{b28} proposed two evasion attacks based on reinforcement learning to generate feature-space AEs. Liu et al.~\cite{b45} used a genetic algorithm to create feature-space AEs for improving the robustness of AMD. Chen et al.~\cite{b43,b44} explored different feature-space evasion attacks (e.g., anonymous attacks and well-crafted attacks) to bypass AMD. Demontis et al.~\cite{b19} proposed a feature-space evasion attack to generate Android AEs by changing the features that seem important for the SVM classifier. Grosse et al.~\cite{b29} generated AEs by modifying the features extracted from Manifest files of Android malware apps using a forward derivative approach. However, the above adversarial attacks might not always yield realizable AEs. In other words, AEs generated by these evasion attacks could be impossible, as they are created by only adhering to norm-bounded constraints without ensuring domain constraints i.e., available-transformations, preserved-semantic, robustness-to-preprocessing, and plausibility constraints~\cite{b17}. In particular, the AEs discussed in~\cite{b19,b20,b21,b28,b29,b44,b43,b45,b46,b62,xu2023gendroid} might not satisfy the robustness-to-preprocessing constraint because the proposed attacks considered adding features to Manifest files in order to generate AEs, while pre-processing operators can discard the added unused features~\cite{b17}. Furthermore, the preserved semantic or plausibility constraints have not been thoroughly investigated in the studies mentioned above. For instance, although the authors in~\cite{b21} tried to manipulate Android malware apps using feature-space perturbations while preserving the malicious functionality of the original apps, they failed to generate realizable AEs since most manipulated apps did not work.

\subsection{Problem-Space AEs}
\label{subsec:problem_space_AEs}
To address the limitation of feature-space norm-bounded AEs,
several studies~\cite{b1,b48,b50,b17,b30,b49} have proposed different approaches for generating AEs.
Specifically, they rely on problem-space transformations that satisfy domain constraints. Labaca-Castro et al.~\cite{b1} generated realizable universal adversarial perturbations by applying a sequence of transformations, found by a greedy algorithm, into malware objects. Bostani and Moonsamy~\cite{b48} proposed an evasion attack that gradually converts an Android malware app into an AE by leveraging transformations identified through querying the target malware detector. Cara et al.~\cite{b50} crafted adversarial Android malware by injecting API calls into malware apps. Pierazzi et al.~\cite{b17} proposed an evasion attack to generate real-world adversarial Android apps through problem-space transformations guided by feature-space perturbations. Chen et al.~\cite{b49} used CW~\cite{b75} and JSMA~\cite{b76} techniques to propose an attack that can mislead AMD. Yang et al.~\cite{b30} introduced two attacks named evolution and confusion attacks to present an Android evasion attack that was based on manipulating Android malware apps. Demontis et al.~\cite{b19} used obfuscation to manipulate Android malware apps. It is worth noting that in the studies above, the problem-space transformations are either code transplantation (incl. harvesting slices of bytecodes extracted from benign apps)~\cite{b1,b17,b30,b48}, obfuscation tools~\cite{b19}, or dummy codes (e.g., unused API calls in Android apps)~\cite{b49}.

However, the practicality of utilizing problem-space transformations in adversarial hardening is debatable due to their high computational complexity~\cite{b39}.
It is also known to be difficult to collect diverse problem-space transformations that fully satisfy the domain constraints~\cite{b48}. For instance, although Yang et al.~\cite{b30} explored domain constraints in the problem space, they failed to generate realizable AEs, as their problem-space transformations caused the apps to crash, mainly because most malware apps cannot run after manipulation. Moreover, Demontis et al.~\cite{b19} utilized \textit{DexGuard}, an Android obfuscation tool, to tamper with malware apps; however, the generated AEs were unable to significantly evade the target detectors they examined, as the obfuscation techniques provided by the tool (i.e., available problem-space transformations) had a limited effect on the features critical for their detectors.

\subsection{Feature-Space realizable AEs}
\label{subsec:feature_space_RealAEs}
Although some studies~\cite{apruzzese2022spacephish,apruzzese2022modeling} argue that \textit{impossible} perturbations can still be employed to generate valid problem-space AEs if adversaries have access to the data-processing pipeline of target systems, there have been several efforts across various domains aimed at overcoming the limitations of both feature-space norm-bounded AEs and problem-space realizable AEs by generating feature-space realizable AEs. Simonetto et al.~\cite{b39} introduced a generic constraint language to define feature dependencies for botnet and credit risk detection. Erdemir et al.~\cite{b38} improved the adversarial robustness of DNN-based models used for malware, spam, and credit risk detection by using non-uniformed perturbations based on the PGD attack~\cite{b33}. Sheatsley et al.~\cite{b51} presented a formal logic framework to learn domain constraints from data used in Network Intrusion Detection Systems (NIDSs) and phishing detection. Teuffenbach et al.~\cite{b68} employed domain knowledge to group flow-based features in NIDSs. Sheatsley et al.~\cite{b40} proposed a domain-constraints-learning method for NIDSs based on independent features that affect other features. Chernikova et al.~\cite{b70} used domain-specific dependencies (e.g., range of feature values) and mathematical feature dependencies to guarantee the realizability of AEs in NIDSs, botnet detection, and malicious domain classification. Tong et al.~\cite{b69} considered so-called conserved features for improving the robustness of PDF malware detection. 

\noindent\textbf{Our work.} To the best of our knowledge, we are the first aiming to not only thoroughly interpret how Android domain constraints (e.g., executability and plausibility of apps) are represented in the feature space for AMD but also propose a technique for learning and applying them.
\begin{figure*}[!t]
    \centering
    \includegraphics[width=1.0\textwidth]{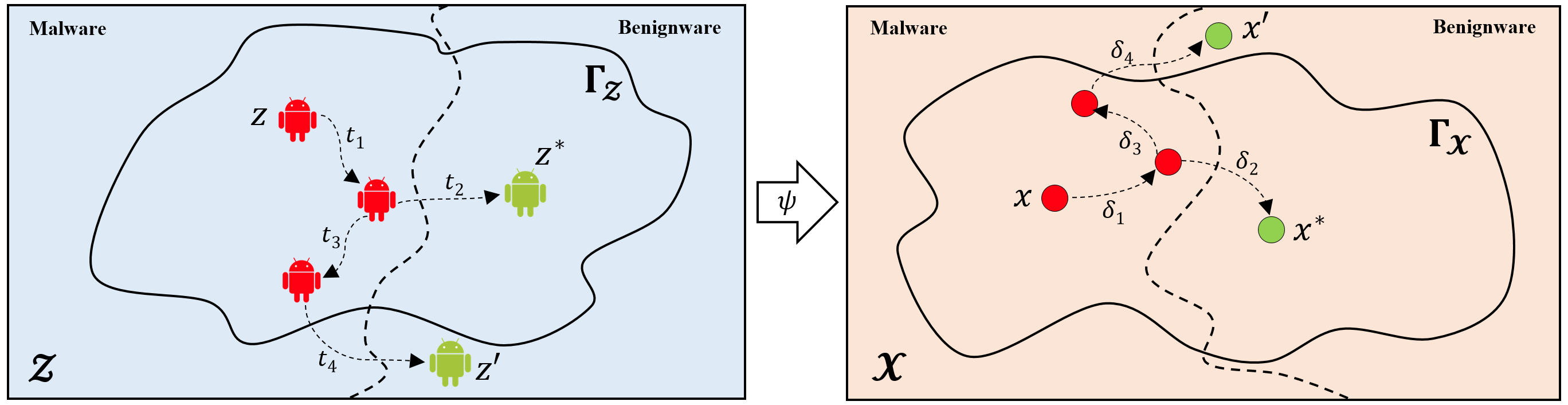}
    \caption{Illustration of generating AEs in the problem space $\set Z$ and the feature space $\set X$ where $\psi$ shows a mapping function from $\set Z$ to $\set X$. The feature-space perturbations $\delta_1$, $\delta_2$, $\delta_3$, and $\delta_4$ correspond to the problem-space transformations $t_1$, $t_2$, $t_3$, and $t_4$, respectively. The dashed lines are the decision boundaries that distinguish malware from benignware. The areas surrounded by solid closed curves represent the realizable problem space and feature space, which meet problem-space domain constraints $\Gamma_{\set Z}$ and feature-space domain constraints $\Gamma_{\set X}$, respectively. $z^*$ and $x^*$ are realizable AEs but $z'$ and $x'$ are unrealizable AEs}.
    \label{fig:problem_feature_space_schema}
\end{figure*}
\section{Interpreting Domain Constraints in the Feature Space}
\label{section:problem-space_constraints_in-the-seature-space}
In the problem space, the domain constraints of Android malware apps are defined as (i) available transformations, (ii) preserved semantics, (iii) robustness to preprocessing, and (iv) plausibility~\cite{b17}; however, we aim to interpret these constraints into a set of new constraints over the feature values in the feature space. Therefore, this section introduces our novel feature-space interpretation of domain constraints for Android malware apps. Before going into the detailed definitions (\S\ref{subsec:problem-space_constraints_interpretation}), we first provide a mathematical background of realizable AEs in both the problem and feature spaces (\S\ref{subsec:problemfeaturespaces}).

\subsection{Problem-Space and Feature-Space Realizable AEs}
\label{subsec:problemfeaturespaces}

Suppose $\psi:\set Z\rightarrow \set X$ is a mapping function that transforms each Android app in the problem space $\set Z$ into a $d-$dimensional feature vector in the feature space $\set X$.
A malware detector is an ML-based binary classifier $f:\set X\rightarrow \set Y$ with a discriminant function $h:\set X \times \set Y \rightarrow \mathbb{R}$ where \(f(x) = \arg\max_{i\in \set Y} h_i(x)\) determines the label of $x \in \set X$. Specifically, $\set Y = \{0,1\}$ is the label set with $y = 0$ indicating benign labels and $y = 1$ indicating malicious labels. Each element in the feature vector $x \in \set X$ is typically discrete~\cite{b3} such as binary representations~\cite{b9,b28,hou2017hindroid,b11,xu2018deeprefiner,kim2018multimodal}, where $0$ indicates the absence and $1$, the presence of a specific feature.
Generally, AEs can be generated by modifying $z\in \set Z$ through problem-space transformations or modifying $x\in \set X$ through feature-space perturbations.

\noindent\textbf{{Problem-Space Realizable AEs}.} In order to generate realizable AEs in the problem space, the following optimization is solved~\cite{b17}:
\begin{equation} \label{eq_optimization_problem_space}
\begin{aligned}
\arg\min_{\set T} \quad & h_{1}(\psi(z'=\set T(z)))~~~\textrm{s.t.} ~~ \set T(z) \vDash \Gamma_{\set Z},\\
\end{aligned}
\end{equation}
where $\set T$ is a sequence of transformations that satisfy the domain constraints defined in the problem space, $\Gamma_{\set Z}$, such as preserving the malicious functionality of the malware~\cite{b21}. 

\noindent\textbf{{Feature-Space Realizable AEs}.} In the feature space, the following optimization is solved~\cite{b17,b19}:
\begin{equation} \label{eq_optimization_feature_space}
\begin{aligned}
\arg\min_{\delta} \quad & h_{1}(x'=x+ \delta)~~~\textrm{s.t.} ~~ \delta \vDash \Omega,\\
\end{aligned}
\end{equation}
where the perturbation vector $\delta$ must satisfy the domain constraints defined in the feature space, $\Omega$.

Most existing studies on generating feature-space AEs do not consider domain constraints but instead, adopt the naive norm bound~\cite{b17} that can lead to AEs being unrealizable. In the feature space, $x' = x + \delta$ is a realizable AE if there exists \textit{at least one} corresponding malware app $z'$ in the problem space (i.e., $\psi(z')=x'$) that not only bypasses malware detection but also satisfies problem-space constraints $\Gamma_{\set Z}$. Figure~\ref{fig:problem_feature_space_schema} illustrates how adversarial transformations in the problem space make adversarial perturbations in the feature space. Reconstructing $z'$ from $x'$ is not possible since $\psi$, i.e., mapping function from $\set Z$ to $\set X$, is neither \textit{invertible} nor \textit{differentiable}~\cite{b17}. For instance, one of the main challenges in converting $x'$, generated by a gradient-based adversarial attack, to $z'$ arises when attempting to back-propagate the loss gradient through the mapping function that behaves like a non-differentiable layer, particularly in non-numerical domains such as malware detection~\cite{eykholt2023uret}. While the inverse feature mapping problem~\cite{b17} presents a considerable challenge in the malware domain, particularly for attackers, defenders are less impacted by this issue because their primary objective is not to generate real adversarial objects but rather to understand which adversarial perturbations are feasible within the model's decision space (i.e., the feature space). To verify the realizability of $x'$, there is no need to directly reconstruct $z'$ from $x'$ to see if $z'$ meets the domain constraints in the problem space because satisfying the domain constraints in the feature space is sufficient. In other words, $x'$ is realizable if $\delta$ meets the domain constraints in the feature space because they demonstrate Android malware properties in the feature space.
\begin{figure}[!t]
    \centering
    \includegraphics[width=0.5\columnwidth]{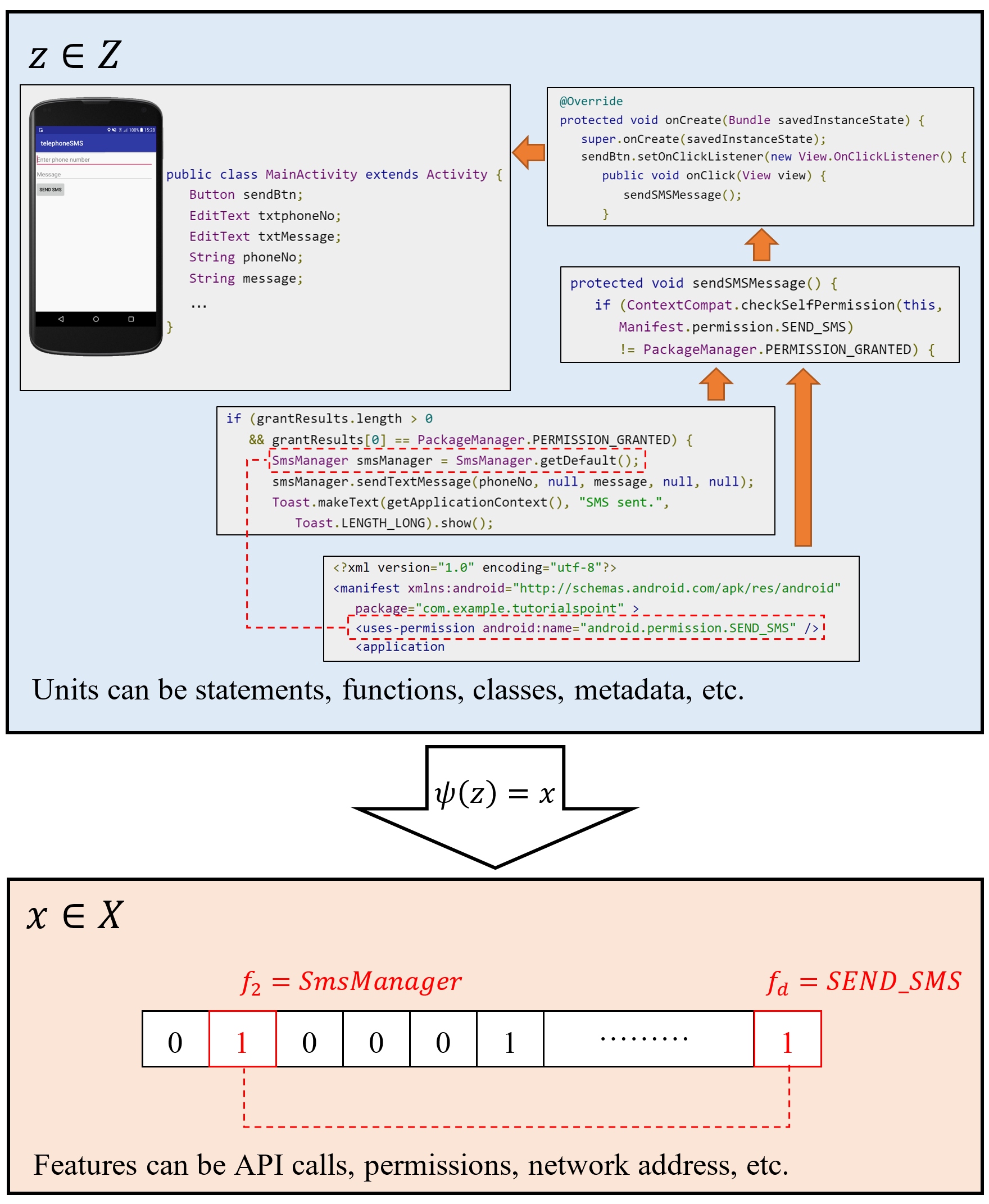}
    \caption{The dependency of two units in the app $z$ is represented by the dependency of two corresponding features in the feature representation $x$.
    } 
    \label{fig:problem_feature_space_dependencies}
    
\end{figure}

\subsection{Domain Constraints in the Feature Space}
\label{subsec:problem-space_constraints_interpretation}
Extracting \emph{all} feature dependencies is not only time-consuming and difficult~\cite{b52} but also unnecessary because this could result in misleading dependencies (i.e., spurious correlations~\cite{quiring2022and}) that have an adverse impact on robustness. Thus, we need to identify the \textit{meaningful} feature dependencies, which can sufficiently guarantee the domain constraints in the feature space. To find out which types of feature dependencies are meaningful, we rely on predefined definitions of domain constraints that have already been formalized within the problem space~\cite{b17}. Specifically, here we introduce our new feature-space interpretation for the four aspects of domain constraints defined in the problem space.
\\

\noindent\textbf{(a) Available perturbations} \textit{refer to all adversarial perturbations $\Delta = \{\delta_1, \delta_2, ..., \delta_n\}$ in the feature space that ensures $x' = x + \delta_i$ meets domain constraints. Using these perturbations makes $x'$ corresponds to \textbf{at least one} problem-space realizable AE $z'$.}\\

Generally, an Android app contains different units (e.g., statements, functions, classes, and metadata) that provide various functionalities.
As shown in Figure~\ref{fig:problem_feature_space_dependencies}, the presence of a specific feature in the feature vector depends on the existence of the corresponding unit in the app. Moreover, the dependencies between multiple units (e.g., \texttt{SmsManager} API and \texttt{SEND\_SMS} permission) also indicate that they offer a particular functionality (e.g., sending messages in Android apps). In the problem space, practical transformations are the ones that consider these sorts of dependencies during app modification. For instance, in the code transplantation technique used to manipulate Android apps~\cite{b1,b17,b30,b48}, an organ (i.e., a problem-space transformation) is extracted from a donor based on the code dependencies because an organ must include all codes associated with a certain functionality~\cite{b85}. In the problem space, the dependencies between units can be clarified by the \textit{System Dependency Graph}~\cite{b85}; however, these dependencies can be extracted from samples in the feature space. Therefore, we argue that using feature dependencies is sufficient for interpreting the domain constraints in the feature space. Specifically, according to the domain constraints defined in the problem space (i.e., preserved-semantic, robustness-to-preprocessing, and plausibility constraints), we introduce the following two sets of dependencies over the feature values in the feature space in (b) and (c).\\

\noindent\textbf{(b) Perfect feature dependencies} \textit{refer to the relationships between pairs of features, signifying that both features in each pair of feature dependencies should occur together. Given a feature-space adversarial example $x' = x + \delta_i$, the perturbation $\delta_i \in \Delta$ might not satisfy domain constraints if $\delta_i$ does not guarantee all \textbf{perfect} feature dependencies.}\\

The semantic equivalence of two programs (e.g., Android apps) is undecidable~\cite{b85}, therefore, in the problem space, adversaries satisfy the preserved-semantics constraint by installing and running the manipulated app $z'$ on an Android emulator and performing smoke testing to make sure that $z'$ can be executed without crashing~\cite{b17,b20,b30,b56}. Similarly, in the feature space, we should ensure that all \textit{perfect} feature dependencies corresponding to an executable app also appear in $x'$.
Otherwise, if only one of the perfectly dependent features exists in $x'$, $z'$ is not a feasible app, as its functionality (e.g., executability) may fail due to the lack of other dependent units.

\noindent\textbf{(c) Relatively strong feature dependencies} \textit{refer to the relationships between each individual feature and the remaining features, highlighting that the feature should appear alongside at least one other feature that frequently occurs with it. Given a feature-space adversarial example $x' = x + \delta_i$, the perturbation $\delta_i \in \Delta$ might not satisfy domain constraints if each feature in $\delta_i$ does not guarantee all \textbf{relatively strong} (including perfect) feature dependencies.}\\

To satisfy the robustness-to-preprocessing constraint in the problem space, it is ensured that preprocessing operators cannot remove unnecessary content (e.g., unused permissions) that has been added to $z$ during generating $z'$~\cite{b17}. Similarly, in the feature space, we ensure that there are no removable added features appearing in $x'$ by keeping the features that have a relatively strong dependency on each added feature. In other words, a specific feature $f_j$ in $x'$ 
is regarded as a removable feature that represents an unused unit $u_j$ in $z'$ when none of its dependent features appears in $x'$.
Moreover, to satisfy the plausibility constraint in the problem space, $z'$ is ensured to be plausible under manual inspection~\cite{b17}. Similarly, in the feature space, we ensure that $x'$ looks plausible when the feature representation is inspected. 

It is important to note that beyond just preserving the semantics, robustness-to-preprocessing, and plausibility constraints further require the adversarial example $z'$/$x'$ to be similar to a realizable app in the problem/feature space.
For this reason, in the feature space, we should ensure that $x'$ keeps all features relatively strongly dependent (i.e., the features that are not necessarily highly dependent, but whose dependencies are stronger than others) in order to achieve a similar feature representation to that of an executable Android app. These dependencies indicate the most dependent features for each feature. Considering relatively strong feature dependencies is sufficient because they capture the most important feature dependencies. Furthermore, feature dependencies in both perfect and relatively strong feature dependencies ensure that dependent features appear together and maintain consistent relationships. This implies that any alteration in one feature must be matched by corresponding changes in its dependent feature to reflect valid and plausible combinations within the feature space. For example, if an adversary changes an app's API level from 30 to 22 while the app still requests permissions relevant only to API level 30, it creates a mismatch and results in unrealistic adversarial perturbations.

\section{Learning Feature-Space Domain Constraints}
\label{section:learning-problem-space-constraints}

This section introduces our learning-based method for extracting the above-defined domain constraints in the feature space. Specifically, we rely on feature correlations to identify perfect feature dependencies, and a graph-based algorithm called Optimum-Path Forest (OPF) to further identify the rest of the relatively strong feature dependencies. Regression analysis is also used in situations where it is necessary to understand how changes in dependent features affect each other. It is noted that OPF is a parameter-independent algorithm~\cite{b55} that essentially considers feature dependencies in our problem to partition dependent features into a cluster.

\noindent\textbf{Preliminaries of Optimum-Path Forest.}
OPF is an efficient pattern recognition algorithm based on graph theory~\cite{b55}. This algorithm reduces a pattern recognition problem to the partitioning of a graph $\set G=(\set V, \set E)$ derived from input dataset~\cite{b53}. $\set G$ is a complete weighted graph wherein the vertices $\set V$ are the feature vectors in the input dataset and the edges $\set E = \set V \times \set V $ are \textit{undirected} arcs that connect vertices. Moreover, each $e_{i,j} \in \set E$ is weighted based on the distance between the feature vectors of corresponding vertices $v_i$ and OPF algorithm works based on a simple hypothesis called \textit{transitive property} in which the vertices belonging to the same partition are connected by a chain of adjacent vertices~\cite{b53}. This algorithm requires several key vertices $\set P \subset \set V $ called \textit{prototypes} that have been found from $\set V$ based on various approaches such as probability density
function~\cite{b59}.
The OPF algorithm partitions $\set G$ into different Optimum-Path Trees (OPTs) where each OPT is rooted at one of the prototypes, through a competitive process among the prototypes to conquer the rest of the vertices~\cite{b53}. In general, the complete weighted graph $\set G$ is partitioned into several OPTs by finding a path from each $v_i \in \set G$ to the best prototype $p \in \set P$, which provides an optimal path with the minimum path cost for $v_i$. 

\begin{figure*}[!t]
    \centering
    \includegraphics[width=0.7\textwidth]{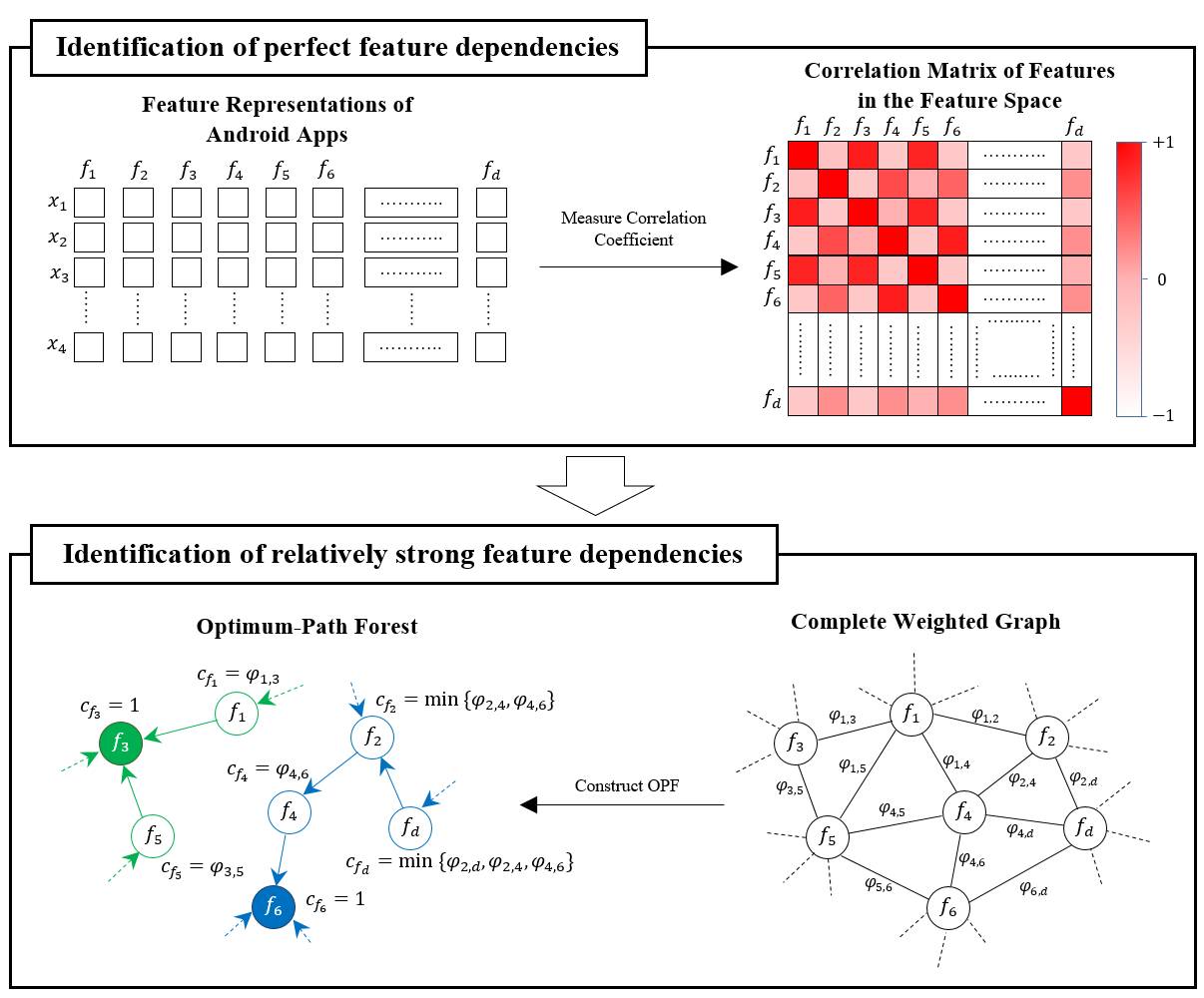}
    \caption{Overview of our method for learning domain constraints from data based on meaningful feature dependencies. $\varphi_{a,b}$ shows \textit{correlation coefficient} between $f_a$ and $f_b$, and $c_{f_a}$ represents the path cost from $f_a$ to the best prototype identified by solving equation~(\ref{eq:cost_function}).}
    \label{fig:block_diagram}
\end{figure*}

\subsection{Our Learning Method}
\label{sec:our-learning-method}
As depicted in Figure~\ref{fig:block_diagram}, our method aims to identify two types of domain constraints $\Gamma'_{\set X} = \{\Upsilon,\Lambda\}$, where $\Upsilon$ and $\Lambda$ show perfect and relatively strong feature dependencies, respectively. In this study, we utilize the correlation coefficient to identify feature dependencies because, given the types of dependencies outlined in~\S\ref{subsec:problem-space_constraints_interpretation}, using this measurement is sufficient for extracting our desired feature dependencies. 

\noindent\textbf{(a) Identification of perfect feature dependencies $\Upsilon$.} 
Based on the types of features being analyzed, a suitable correlation coefficient should be chosen to measure the correlation between every pair of features. The dependency between a pair of features is perfect if the correlation coefficient between them equals 1. We create $\Upsilon$, the set of all perfect feature dependencies based on the identified perfect correlations. Note that each $\set B_i \in \Upsilon$ includes all features in $\set F$ that are perfectly correlated where $\set F$ is the feature set of $\set X$.

\noindent\textbf{(b) Identification of relatively strong feature dependencies $\Lambda$.} Based on our explanation in \S~\ref{subsec:problem-space_constraints_interpretation}, considering only the perfect feature dependencies is insufficient for ensuring that feature-space AEs closely resemble the feature representations of realizable Android apps.
For this reason, we adopt OPF to further learn other relatively strong feature dependencies beyond the perfect ones. The proposed version of OPF partitions $\set F$ into the different groups $\set A_i$ where the features that are more interdependent belong to the same cluster. As shown in Figure~\ref{fig:block_diagram}, to construct OPF, we first create a complete weighted graph $\set G=(\set V, \set E)$ where $\set G = \set F$, and $\set E = \set F \times \set F$ includes the edges between each pair of features $(f_a, f_b)$ weighted by correlation coefficient. Then, from each set of very strongly correlated features (i.e., $\varphi > 0.9$), we randomly select one feature as a prototype. This is due to the fact that highly correlated features can naturally indicate a potential cluster, making them suitable for clustering the remaining features. Finally, $\set G$ is partitioned based on the typical method used in the OPF algorithm which is slightly modified, particularly its connectivity and cost functions, because here, the weights of edges are specified based on the correlation coefficient instead of distance as in the original algorithm. Suppose $\pi_{f_b,f_a}=\langle f_b, ..., f_k,f_a \rangle$ is a path from $f_b$ to $f_a$. In the modified OPF algorithm, a connectivity function $f_{min}$, which is a smooth function, assigns a path cost to each path as follows: 

\begin{equation}
    \begin{split}
   & f_{min}(\langle f_b \rangle) = 
      \begin{cases}
      1 & \text{if $f_b \in \set P$}\\
      -\infty & otherwise
    \end{cases}\\
   & f_{min}(\pi_{f_b,f_k}.\langle f_k , f_a \rangle)= min\{f_{min}(\pi_{f_b,f_k}),\varphi(f_k,f_a\}
   \end{split}
   \label{eq_path_cost_function}
\end{equation}
where $\pi_{f_b,f_k}.\langle f_k , f_a \rangle$ shows the connection of the edge $\langle f_k , f_a \rangle$ to the path $\pi_{f_b,f_k}$. As shown in equation~(\ref{eq_path_cost_function}), the path cost of $\pi_{f_b,f_a}$ is the minimum weight of edges along the path. The modified OPF algorithm aims to find an optimal path for each $f_a \in \set F$ by maximizing $f_{min}$ through the following cost function:

\begin{equation}
     \begin{split}
     Cost(f_a)= \max_{\forall f_p \in \set P, \pi_{f_p,f_a}} \quad & \{ f_{min}(\pi_{f_p,f_a}) \}.
     \end{split}
     \label{eq:cost_function}
\end{equation}
where $\set P$ shows the prototype set. Optimum-path trees constructed by the OPF algorithm let us determine other relatively strong correlations because an OPT includes a subset of all features in the feature space (i.e., $\set A \subset \set F$) where each feature $f_a \in \set A$ is more dependent on other features in $\set A$ as compared to the rest of features $\set F \backslash \set A$. According to the specified OPTs, we create $\Lambda$ which is the set of all relatively strong feature dependencies. Each $\set A_i \in \Lambda$ contains all the features in $\set F$ that are relatively strongly correlated.

Note that we also demonstrate that our OPF-based identification method is better than a simple baseline that uses a fixed threshold to keep highly correlated features -- see results in \S~\ref{sec:evaluate_OPF}. 

\noindent\textbf{Regression Analysis.} 
For non-binary feature spaces used by some detectors like MaMaDroid~\cite{b8}, where its features represent Markov transition probabilities between API calls, we need to not only specify dependent features but also understand how variations in one feature affect others. 
Specifically, to show how adversarial perturbations in feature $f_a$ influence feature $f_b$, we fit a regression model with $f_a$ as the independent variable and $f_b$ as the dependent variable. This model helps predict feasible values of $f_b$ based on $f_a$, ensuring $f_b$ looks representative of feasible apps.
\section{Applying Feature-Space Domain Constraints}
\label{section:applying-the-problem-space-constraints}
This section explores two approaches to demonstrate how our learned domain constraints can be applied to counter evasion attacks.

\subsection{Adversarial Example Detection}
\label{subsection:ae_detection}
According to our learned domain constraints, we propose a technique to identify AEs in advance, before engaging the ML model constructed for AMD. Specifically, we first introduce a way to validate how our learned feature-space domain constraints can represent the domain constraints of feasible
apps. To this end, we define a new metric called \emph{Constraints Satisfaction Rate} (CSR) to measure the ratio of the features that satisfy our learned domain constraints to all the features of a particular sample. By satisfying our learned domain constraints, we mean that one specific feature appears simultaneously with at least one of its relatively strong dependent features and all its perfectly dependent features specified in $\Lambda$ and $\Upsilon$, respectively. Then, we use CSR as a criterion to distinguish AEs from feasible
apps, such that an input app is considered AE if the CSR of its feature representation falls below a threshold. This criterion aids in confronting evasion attacks (e.g., ~\cite{xu2023gendroid,zhang2021shadowdroid,b29}) that generate AEs through norm-bounded perturbations without considering the properties of feasible apps. For further details on these attacks, refer to \S\ref{subsec:feature_space_AEs}.

Note that we do not expect the feature representation of feasible apps to fully satisfy our learned domain constraints because as shown in Figure~\ref{fig:feature_space_domain_constraints}, our learned domain constraints are indeed a subset of true feature-space domain constraints. This is mainly because they are learned from a finite set of samples that might not fully represent the true distribution of all existing apps.

\subsection{Adversarial Hardening}
\label{subsection:adversarial_hardening}
This defense approach emphasizes integrating AEs into the training process of the classifier alongside the original training set~\cite{b42}. In this paper, we introduce two defense techniques based on adversarial hardening, which are proposed by adapting typical adversarial training~\cite{b33} and adversarial retraining~\cite{b54} approaches to employ our learned domain constraints.
\subsubsection{Adversarial Training with Domain Constraints}
\label{section:adversarial_training}
\begin{figure}[!t]
    \centering
    \includegraphics[width=0.5\columnwidth]{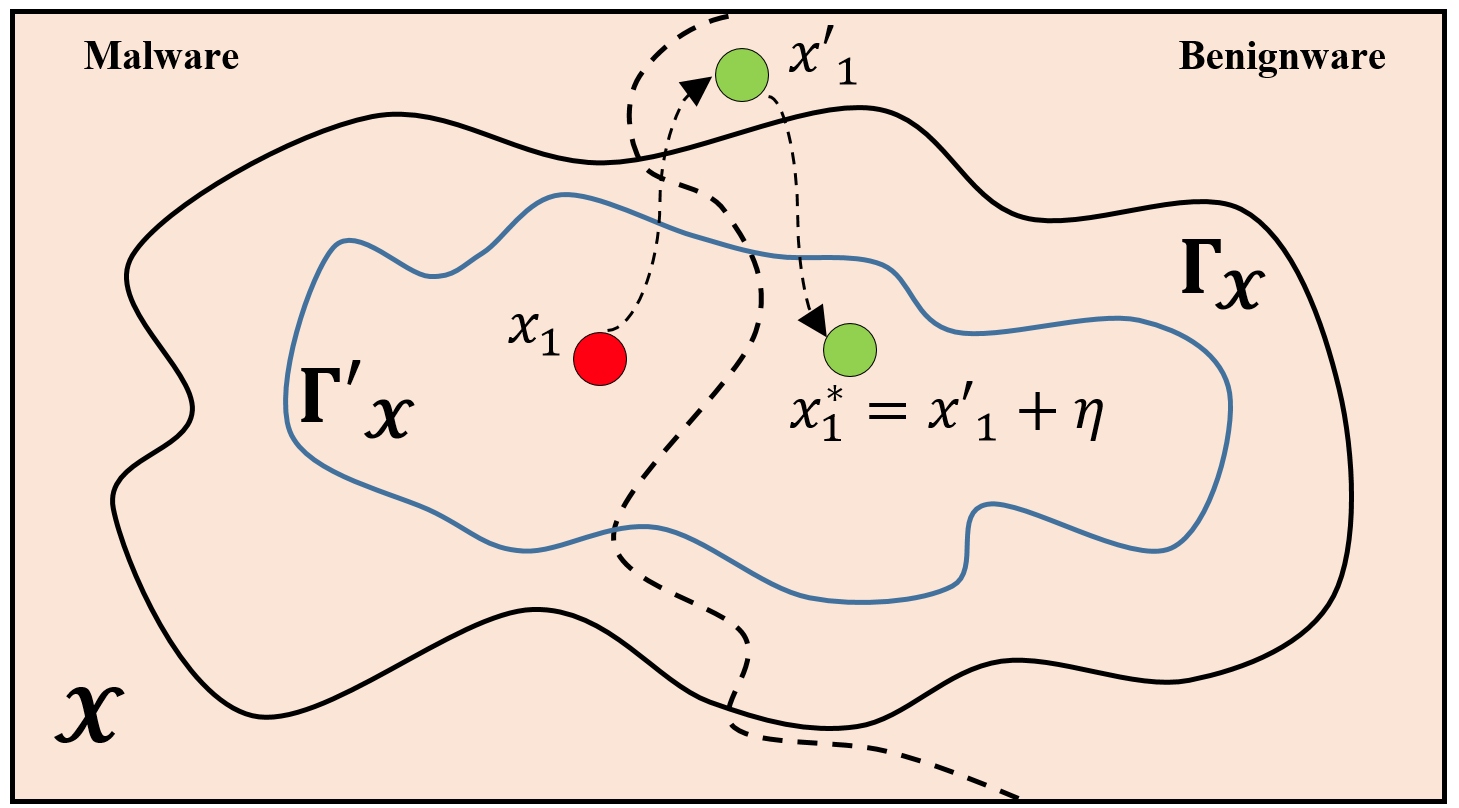}
    \caption{Illustration of generating a feature-space realizable AE $x^*_1$ by adding missed meaningful dependent features $\eta$ to unrealizable AE $x'_1$. The area surrounded by the black closed curve represents the actual realizable feature space determined by the complete domain constraints $\Gamma_{\set X}$, while the blue closed curve area represents the realizable feature space determined by our learned domain constraints $\Gamma'_{\set X}$. Our learned realizable space is a subset of the actual realizable space due to the limitation of learning from finite data.}
    \label{fig:feature_space_domain_constraints}
\end{figure}

AT is a well-established defense strategy against AEs that is widely used in the context of Android malware~\cite{b1,b20,b21,b28,b29,b30}. This defense strategy proactively incorporates the generation of AEs into the training phase of ML models~\cite{b23}. It solves the following min-max optimization for AT~\cite{b42,zeng2021adversarial}:

\begin{equation}
     \begin{split}
     \min_{\theta} \mathbb{E}_{(x_i,y_i) \sim \set D}\quad & [ \max_{\delta \vDash \{\Omega, \Gamma'_{\set X}\}} \set{L}(f_{\theta}(x+\delta),y) ]
     \end{split}
     \label{eq:cost_function_new}
\end{equation}

\noindent where $\set{L}$ denotes the loss function and $\theta$ denotes the parameters of the Android malware detector $f_{\theta}$. Moreover, $\mathbb{E}$ is the expected value of inner optimization according to $(x_i,y_i) \sim \set D$ indicating training data samples drawn from the distribution $\set D$. As shown in equation~\ref{eq:cost_function_new}, the adversarial perturbations $\delta$, which is found by solving the inner optimization, must satisfy not only initial feature space constraints (i.e., $\delta \vDash \Omega$), which is often norm-bounded constraints but also our learned feature-space domain constraints (i.e., $\delta \vDash \Gamma'_{\set X}$) because as depicted in Figure~\ref{fig:feature_space_domain_constraints}, satisfying $\Gamma'_{\set X}$ can turn an unrealizable AE into a realizable AE. Lines 2 to 13 in Algorithm~\ref{algorithm:generate_ae} show how an adversarial perturbation $\delta$ becomes realizable by adding dependent features. Note that in Line~10 of Algorithm~\ref{algorithm:generate_ae}, we select a feature from $\set A_i \backslash \{f_j\}$ into $\delta^*$ that seems the most important for the input attack $\mathbf{Att.}$, e.g., the feature associated with the highest gradient provided by a gradient-based attack. In Lines 5 (and 11), we construct a regression model $\mathcal{M}$ using the training set, with $f_j$ as the independent variable and $f_b$ (and $f_a$) as the dependent variable. $\mathcal{M}$ is used to estimate plausible values for $f_b$ (and $f_a$) based on $f_j$. To minimize the computational overhead of Algorithm~\ref{algorithm:generate_ae}, we cache new regression models, eliminating the need to rebuild them for similar cases. It is important to note that regression models are employed for estimating dependent feature values only when the input feature space is non-binary. For a binary feature space, it is sufficient just to add the dependent features (i.e., skip Lines 5 and 11 in Algorithm~\ref{algorithm:generate_ae} and adjust Lines 6 and 12 to include $f_b$ or $f_a$ in $\delta^*$).

\begin{algorithm}[ht]
\SetAlgoLined
\KwIn{$\mathbf{Att.}$: a feature-space adversarial attack;

$\delta$: an adversarial perturbation found by $\mathbf{Att.}$; 

$\Gamma'_{\set X} = \{ \Upsilon,\Lambda \}$: feature-space domain constraints.}
\KwOut{$\delta^*$, a realizable adversarial perturbation.}
  $\delta^* \leftarrow \delta$.\\
  \ForEach{feature $f_j$ in $\delta$}
  {
        \If{there exists $\set B_k \in \Upsilon$ including $f_j$}{                      
            \ForEach{feature $f_b \in \set B_K$}
            {
                Load the regression model \(\mathcal{M}\) for \( f_j \rightarrow f_b \) if it exists; otherwise, build it.\\
                Add $\set M(f_b)$ into $\delta^*$.
            }            
        }    
        Find $\set A_i \in \Lambda$ containing $f_j$\\
        Select $f_a \in \set A_i \backslash \{f_j\}$, which appears the most important for $\mathbf{Att.}$\\
         Load the regression model \(\mathcal{M}\) for \( f_j \rightarrow f_a \) if it exists; otherwise, build it.\\
         Add $\set M(f_a)$ into $\delta^*$.\\  }
  \Return $\delta^*$
 \caption{Applying our feature-space domain constraints}
 \label{algorithm:generate_ae}
\end{algorithm}

\begin{algorithm}[t]
\SetAlgoLined
\KwIn{$(x,y)$: a malware sample where $x$ and $y$ is the feature vector of the sample and its label; $f_{\theta}$: target malware detector with parameters $\theta$; $\set L$: loss function of $f_{\theta}$; $k$: steps; $\alpha$: step-size; $q$: percentile; $\epsilon$: the $L_1$ norm perturbation bound; $\Gamma'_{\set X}$: our learned domain constraints; $\mathcal{F}$: the feature set characterizing the dimensions of feature space $\mathcal{X}$}.
\KwOut{ $x'$, the perturbed samples.}
  $\delta \leftarrow$ a vector of 0 with length $x$.\\ 
  \ForEach{$i = 1$ to $k$}
  {
        $g \leftarrow \nabla_{\delta_i}\set L(f_{\theta}(x+\delta_i),y)$.\\       
        $e \leftarrow \{r(g_j)$ \textbf{if} $ g_j \geq P_q(g)$ \textbf{else} $0 | 1 \leq j \leq d\}$ where $r(g_j)$ is a function that rounds the value $g_j$ up to the nearest integer, $d$ is the dimension of $x$ and $P_q(g)$ is the $q$-th percentile of $g$.\\
        $\delta_i \leftarrow \delta_{i-1}+\alpha \times e$.\\
        $\delta_i \leftarrow \epsilon.\frac{\delta_i}{\max\{\epsilon,\lVert x \rVert_1\}}$.\\       
  }
  \If{$\mathcal{F}$ is composed of discrete features}{
      $C, U \leftarrow$ Select top-$\epsilon$ features in $\delta$ with highest values, and their corresponding values.\\      
      $\delta, \delta' \leftarrow $ two vectors of 0 with length $x$.\\          
      \While{$\lVert \delta \rVert_1\ \leq \epsilon$}{
            $c \leftarrow$ Remove top-1 feature from $C$.\\            
            $\delta_c' \leftarrow U[c]$.\\ 
            Apply Algorithm~\ref{algorithm:generate_ae} to ensure $\delta'$ satisfies $\Gamma'_{\set X}$.\\
            \If{$\lVert \delta \rVert_1\ + \lVert \delta' \rVert_1\ \leq \epsilon$}{
              $\delta \leftarrow \delta + \delta'$.\         
            }          
            $\delta' \leftarrow $ a vector of 0 with length $x$.\\
      }
  }
  
  $x' \leftarrow x + \delta$.\\
  {\Return $x'$} 
 \caption{PGD Attack under $L_1$ bounds and our feature-space domain constraints}
 \label{algorithm:pgd}
\end{algorithm}

Over the past years, Projected Gradient Descent (PGD)~\cite{b33} has been extensively applied in the field of malware detection~\cite{b20,b21,liu2023a2,li2023pad}. This paper uses PGD adapted for the Android malware domain~\cite{b21} to find $\delta$. We adopt the $L_1$ norm bound as the perturbation bound for PGD. This attack, which is described in Algorithm~\ref{algorithm:pgd}, has been modified slightly for the purpose of our study. Specifically, we follow the suggestion in~\cite{b21} and incorporate a normalization step to address the small-gradients issue, which may occur especially when the feature space is binary. This step involves updating the perturbation $\delta$ in the steepest gradient direction, which is computed as the unit vector $e$ with $e_{j^*} = sign(g_{j^*} )$ for $j^*=\arg\max_{1 \leq j \leq d}|g_j|$~\cite{tramer2019adversarial}, where $g_j$ is the value of the $j$-th index of the gradient $g= \nabla_{\delta_i}\set L(f_{\theta}(x+\delta_i),y)$ computed in $i$-th iteration of PGD, and $d$ is the dimension of sample $x$. In our study, we consider $g_j$ rather than  $|g_j|$ since our attacker can only add features for generating AEs. Moreover, as shown in Line 4 of Algorithm~\ref{algorithm:pgd}, we adopt the proposed solution from~\cite{tramer2019adversarial} to address the inefficiency of updating a single feature by updating multiple features simultaneously. We use the projection operator demonstrated in~\cite{b38} to project perturbation $\delta_i$ into $L_1$ norm bound with the size of $\epsilon$ (Line 6 in Algorithm~\ref{algorithm:pgd}).
Note that as stated in~\cite{b33}, this attack lets $\delta$ be continuous during the optimization process. However, as shown in Lines 9 to 19, if the input feature space is discrete, we map $\delta$ to the 
discrete feature space by considering at most $\epsilon$ indices in $\delta$ (including the top features with the highest values and their dependent features specified by Algorithm~\ref{algorithm:generate_ae}) before incorporating $\delta$ into the input malware sample. We ensure that the process of adding each feature, including its dependent features, from the top-$\epsilon$ features of the initially identified $\delta$ to the final $\delta$ is carried out in a manner that respects the perturbation bound $\epsilon$ unless the feature should not be included. 
Our mapping approach makes $\delta$ a realizable adversarial perturbation; however, considering only the top-$\epsilon$ indices in $\delta$ might lead to creating an unrealizable adversarial perturbation.

It is important to note that besides PGD, various other methods, especially those designed for discrete feature spaces~\cite{yang2020greedy,gao2023discrete}, can be used in the inner maximization problem of equation~\ref{eq:cost_function_new} to identify adversarial perturbations. In Appendix~\ref{apendix:sparse_rs}, we examine Sparse-RS~\cite{b62}, which was originally tested on AMD, instead of PGD, and evaluate its performance. Moreover, since we are focused on defense, the goal of Algorithm~\ref{algorithm:generate_ae} is not to generate realizable AEs that mislead malware classifiers but rather to convert adversarial perturbations produced during robust optimization into realizable ones. Although incorporating dependent features in these perturbations may reduce misclassification confidence compared to the original adversarial perturbations, our preliminary analysis demonstrates that the misclassification confidence of the feature representations of malware samples adversarially modified by our approach is still significantly higher than that of the original malware samples, showcasing their effectiveness in AT.

\subsubsection{Adversarial Retraining with Domain Constraints}
\label{section:adversarial-retraining}
This adversarial hardening method directly uses AEs to augment the training data. Suppose $\set D_m = \{(x_i,y_i)|x_i \in \set X, y_i=1, i = 1, ..., k\}$ shows a fraction of all the malware samples in the input training set $\set D$. In adversarial retraining, we first construct an adversarial set $\set D^a_m = \{(x'_i,y_i)|x_i \in \set X, y_i=1, i = 1, ..., k\}$ where each $x'_i$ is the adversarial example of $x_i$ that is generated by using an evasion attack. Then, we use $\set D' = \set D \cup D^a_m$, the mixed set that is augmented with AEs, to retrain the ML models.

In this study, we consider our learned domain constraints in the evasion attack that is supposed to prepare $\set D^a_m$ because we aim to augment $\set D$ with realizable AEs. In fact, every adversarial perturbation $\delta$ generated by the evasion attack must satisfy both the feature-space constraints and our learned domain constraints (i.e., $\delta \vDash \Omega$ and $\delta \vDash \Gamma'_{\set X}$). Note that to adhere to our learned domain constraints, the evasion attack should add an adversarial feature alongside one of its relatively strong dependent features, as well as all its perfectly dependent features specified in $\Lambda$ and $\Upsilon$, respectively. The values of the newly added dependent features are determined based on the adversarial features using a regression model $\mathcal{M}$, similar to the method described in Algorithm~\ref{algorithm:pgd}.
\section{Experimental Results}
\label{section:simulation-results}
In this section, we empirically evaluate the performance of our learned domain constraints in confronting evasion attacks aimed at tricking AMD. Specifically, our experiments aim to answer the following research questions:

\begin{itemize}    
     \item[\textbf{\hypertarget{RQ1}{RQ1}.}]    
    Can our learned domain constraints be applied to counter evasion attacks that generate AEs without ensuring their feasibility?   
    (\S\ref{section:evaluate-realizable-feature-space-perturbations-in-adversarial-retraining})
    \item[\textbf{\hypertarget{RQ2}{RQ2}.}] 
    Can our learned domain constraints help AT enhance the robustness of the detectors against realistic evasion attacks?
    (\S\ref{section:AT-evaluation})
    \item[\textbf{\hypertarget{RQ3}{RQ3}.}] Can our OPF-based method effectively extract meaningful feature dependencies?(\S\ref{sec:evaluate_OPF})
    \item[\textbf{\hypertarget{RQ4}{RQ4}.}] Can our feature-space realizable AEs outperform the conventional problem-space realizable AEs?(\S\ref{sec:feature_vs_problem_space_comparison})
\end{itemize}

All the experiments have been performed on a Debian Linux workstation with an Intel (R) Core (TM) i7-4770K, CPU 3.50 GHz, 32 GB RAM, and GPU GeForce RTX 3080 Ti.

\subsection{Experimental Setup}
\noindent\textbf{Threat Models and Attacks.}
The evasion attacks considered in our experiments generate AEs based on the threat model that is described with three attributes:
\begin{itemize}
    \item \textbf{Adversary's Goal.} The goal of the adversary is to trigger the Android malware detector to misclassify the adversarial (malware) example as benign.
    
    \item \textbf{Adversary's Knowledge.} The adversary may have perfect knowledge (PK), limited knowledge (LK), or zero knowledge (ZK) about the target model, including its learning algorithm, training data, feature space, and parameters. In other words, in PK, LK, and ZK attacks, the target model is considered as a white-, gray-, and black-box model by the adversary, respectively. In our experiments, we assess the efficacy of our approach using whole three types of attacks.
    
    \item \textbf{Adversary's Capability.} The adversary can generate AEs either in the feature space by perturbing feature representations under feature-space constraints, or in the problem space by applying a sequence of transformations under domain constraints~\cite{b17}. Here we follow the common practice of only considering feature-addition transformations/perturbations~\cite{b17,b29,b48}.
\end{itemize}

To explore \textbf{\hyperlink{RQ1}{RQ1}}, we use three evasion attacks, GenDroid~\cite{xu2023gendroid}, ShadowDroid~\cite{zhang2021shadowdroid}, and Grosse attack~\cite{b29}, which generate AEs under ZK, LK, and PK settings, respectively. These attacks generate adversarial perturbations irrespective of domain constraints, raising doubts about the realizability of the resulting AEs. Moreover, our work considers a realistic problem-space attack, known as \textit{PiAttack}~\cite{b17}, to empirically investigate different RQs stated in \S\ref{section:simulation-results}. PiAttack is a white-box attack that generates problem-space realizable AEs by applying effective problem-space transformations (i.e., code snippets called gadgets extracted from donor apps) specified by feature-space perturbations. Note that PiAttack can be regarded as an adaptive attack, as it inherently knows our domain constraints. We refer the reader to Appendix~\ref{apendix:PiAttack} for the technical details of PiAttack.

As a comparison, we also consider the well-known PGD attack introduced in \S\ref{section:applying-the-problem-space-constraints}, which directly adds perturbations in the feature space constrained by specific $L_1$ norm bounds.
Specifically, we use PGD for both AT and attacking target detectors. Furthermore, in the experiments involving adversarial retraining (i.e., \S\ref{sec:evaluate_OPF} and \S\ref{sec:feature_vs_problem_space_comparison}), we have developed a feature-space attack called PK-Feature, which operates in a manner similar to PiAttack within the feature space. See Appendix~\ref{appendix:pk_feature} for more details about this attack.

\noindent\textbf{Target Detector.}
We use the well-known DREBIN-Support Vector Machine (SVM)~\cite{b9} as a baseline target Android malware detector. This detector builds a linear SVM in a binary feature space consisting of eight types of features (e.g., permissions and restricted API calls). The regularization hyperparameter of the implemented linear SVM is $C=1$~\cite{b17}. It is worth mentioning that for AT, we also consider DREBIN-Deep Neural Network (DNN)~\cite{b1} consisting of four layers with the dimensions as $10,000 \times 1,024 \times 512 \times 2$. We also take into account another DNN-based malware detector called DroidAPIMiner-DNN. This malware detector, which leverages the DroidAPIMiner~\cite{b11} feature representation (i.e., a binary feature representation solely constructed from API calls appearing in Android apps), consists of four layers with the dimensions as $337 \times 256 \times 128 \times 2$, where $337$ represents the dimension of our samples represented based on DroidAPIMiner feature representation. Given the potential limitations in DREBIN and DroidAPIMiner, as discussed in~\cite{daoudi2022deep} and~\cite{onwuzurike2018family} respectively, we consider a more advanced malware detector called RAMDA-DNN. This new malware detector, built upon a robust feature representation derived through Autoencoder~\cite{li2021robust}, exhibits an architecture similar to DroidAPIMiner but instead uses 269 in the input layer, showing the dimensions of samples. 
All DREBIN-DNN, DroidAPIMiner-DNN, and RAMDA-DNN employ ReLU activation functions in their hidden layers and a Sigmoid activation function in their output layer. We train these malware classifiers for $100$ epochs with a batch size of $1024$.

While binary feature representation is common in the malware domain, we consider another family of malware detectors called R-PackDroid-DNN, which operates on a discrete, non-binary feature representation to demonstrate the broad applicability of our method. This detector utilizes the R-PackDroid-DNN~\cite{scalas2019effectiveness} feature representation, where each feature represents the frequency of a specific type of system API component, such as a package, class, or method, within Android apps. The R-PackDroid-DNN model is composed of four layers with dimensions $2155 \times 1024 \times 512 \times 2$, where $2155$ corresponds to the dimensionality of the samples, based on the R-PackDroid feature representation derived from the classes of system API components employed in the apps.

\noindent\textbf{Dataset.}
We use a public Android dataset~\cite{b17} including $\approx152K$ Android apps collected from AndroZoo~\cite{b67}.
In this dataset, an app is defined as malware if it is detected by $4+$ VirusTotal AVs, and as a benign sample, if no AVs detect it.
We randomly select a test set of $30K$ samples, comprising $25K$ benign and $5K$ malware samples, while the remaining $\approx122K$ samples, including $\approx111K$ benign and $\approx11K$ malware samples, are designated as the training set. Note that we consider a fair proportional distribution between benign and malware samples to mitigate spatial bias~\cite{pendlebury2019tesseract}. To assess DREBIN, samples are represented based on the DREBIN feature representation~\cite{b9}, a widely used binary feature set in recent studies~\cite{b1,b17,b21,b46,b48,b62}.
Since the DREBIN feature space is a very high-dimensional (i.e., over $1M$ features) but sparse feature space, with a significant amount of redundant features impacting DREBIN's performance~\cite{daoudi2022deep}, we select the $10K$ most distinguishing features following the recommendations from previous studies~\cite{b19,b20}. We perform feature selection and feature dependency extraction solely on the training set, preventing the data-snooping pitfall~\cite{quiring2022and}. To measure the correlation between every pair of features, which is crucial for extracting meaningful feature dependencies, we use \textit{phi coefficient}~\cite{b58} because the feature space of target detectors considered in our evaluation consists of binary features.
To evaluate the adversarial robustness of different Android malware detectors, we randomly select $1K$ malware samples from the test set, representing diverse malware families, to generate the AEs.

\noindent\textbf{Evaluation Metrics.}
For evaluating the malware detectors, we consider Accuracy (Acc) as well as True Positive Rate (TPR) and False Positive Rate (TPR). Specifically, we calculate \emph{clean Acc} on benign and malware examples for model utility and \emph{robust Acc} on adversarial malware examples for robustness.

\begin{figure}[!t]
    \centering
    \includegraphics[width=0.5\columnwidth]{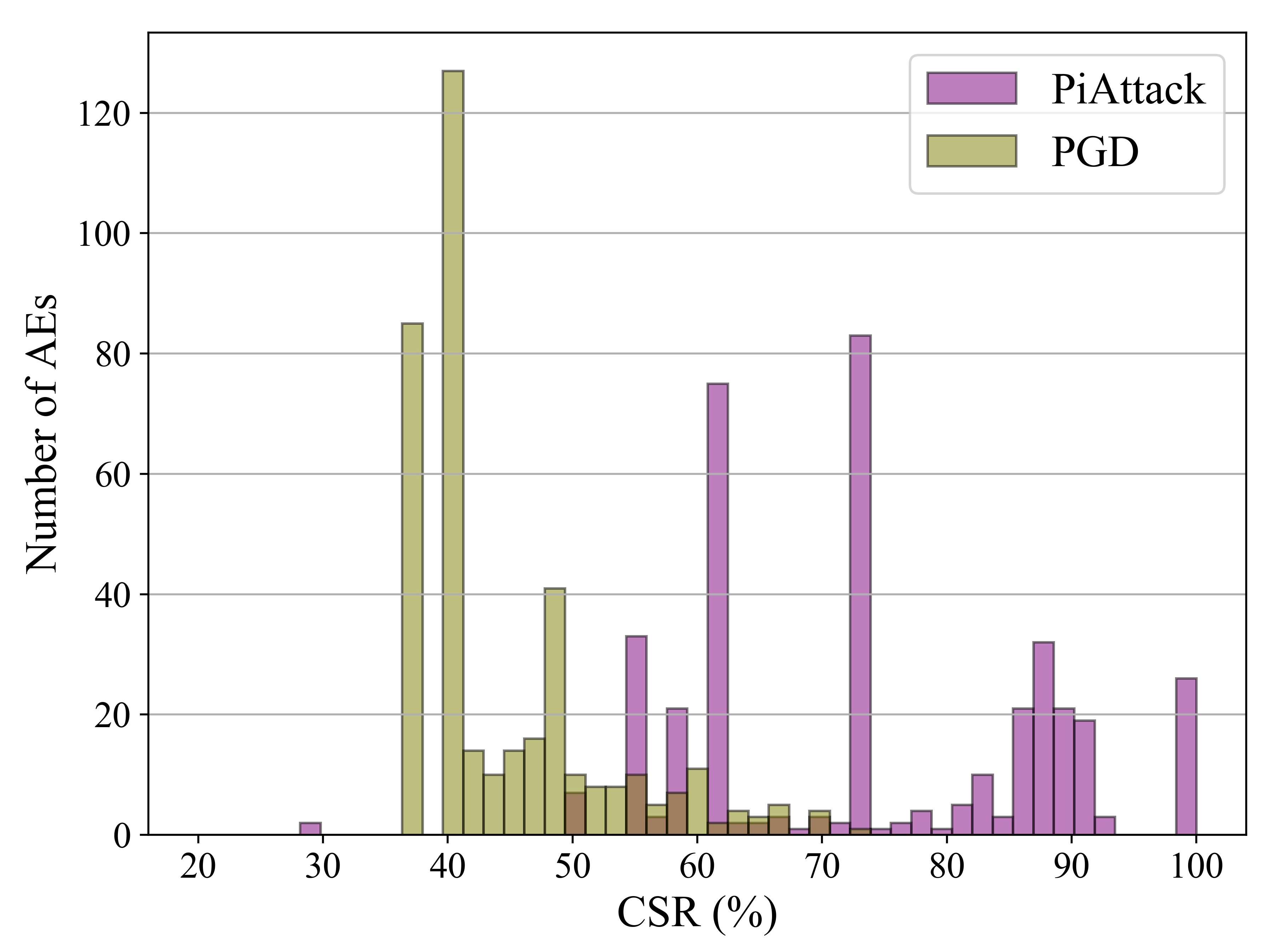}
    \caption{CSR of AEs generated by the domain-constraint-aware attacks, PiAttack vs. the domain-constraint-agnostic attacks, PGD, when they target DREBIN-DNN.}
    \label{fig:plot_compare_apps_problem_space_satisfaction}
\end{figure}

\subsection{Evaluating Our Learned Domain Constraints}
\label{section:evaluate-realizable-feature-space-perturbations-in-adversarial-retraining}
To answer \textbf{\hyperlink{RQ1}{RQ1}}, we first validate the utility of our learned domain constraints for representing Android malware properties by empirically evaluating if they can help to distinguish realizable AEs from unrealizable AEs. To this end, we demonstrate how the added features in AEs generated by different attacks on the DREBIN-DNN, which is based on standard training, can satisfy our learned domain constraints.
Specifically, we consider two attacks, i.e., the domain-constraint-aware attack, PiAttack, and the domain-constraint-agnostic attack, PGD.
Both attacks are bounded by the $L_1$ norm $\epsilon=30$.
We calculate CSR defined in \S\ref{subsection:ae_detection} for the AEs successfully generated by both attacks. Figure~\ref{fig:plot_compare_apps_problem_space_satisfaction} demonstrates that AEs generated by the realistic attack can better satisfy our learned domain constraints than the unrealistic attack.
Specifically, the average CSR of AEs generated by PiAttack is $73.8\%$, and that of the AEs generated by PGD is only $44.6\%$. The relatively high CSR results related to PiAttack confirm that our extracted feature dependencies can adequately represent the domain constraints. Note that we expect that using a larger number of training samples would further improve the results.

\begin{figure}[!t]
    \centering
    \includegraphics[width=0.5\columnwidth]{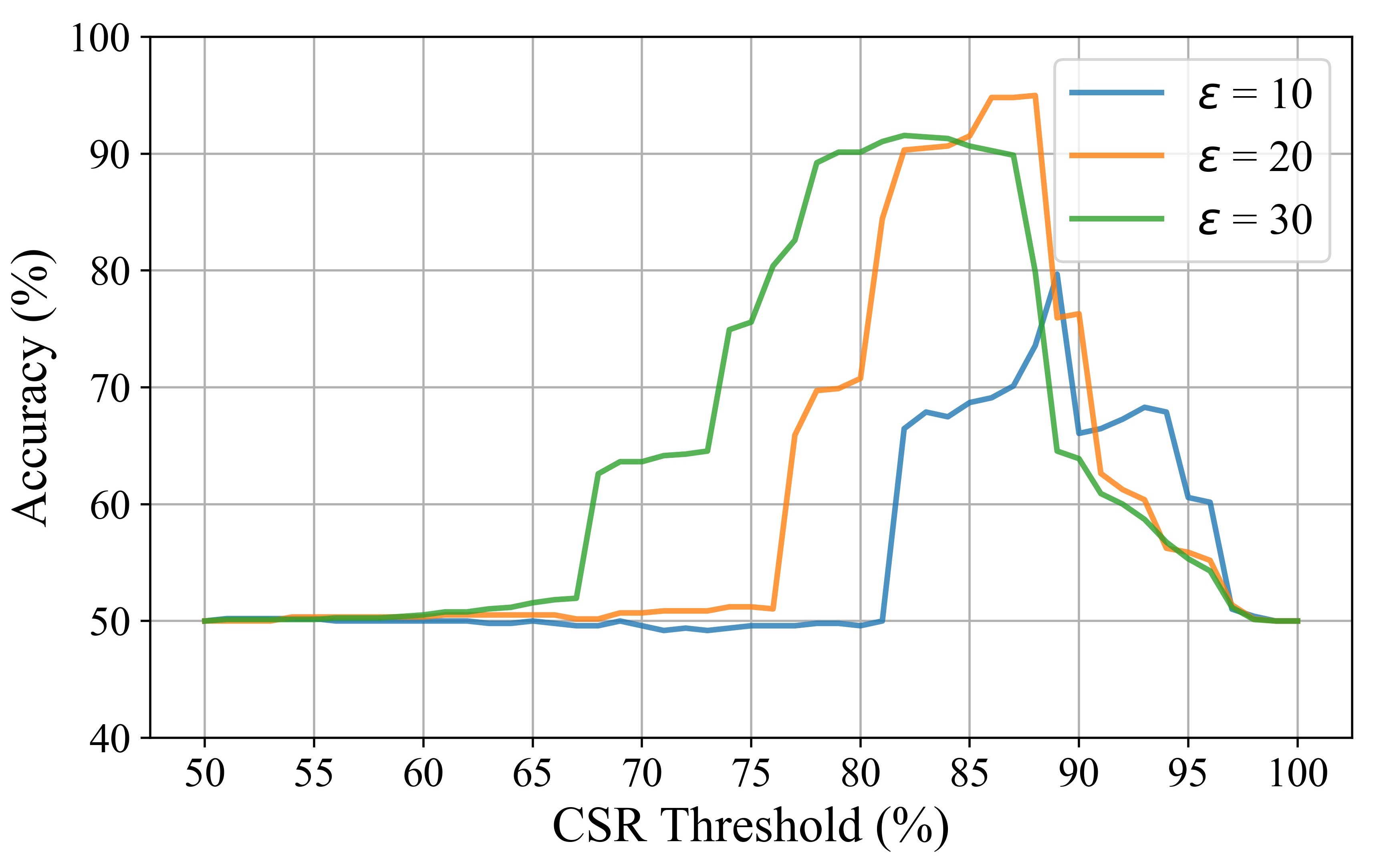}
    \caption{Differentiating realizable AEs generated by PiAttack from unrealizable AEs generated by PGD based on our learned domain constraints when the attacks target DREBIN-DNN. Results are reported for various $L_1$ norm bounds.}
    \label{fig:acc_differentiation}
\end{figure}

Our empirical evaluation suggests that it is possible to differentiate realizable AEs from unrealizable AEs by setting a CSR threshold.
Specifically, we consider the AEs above this CSR threshold as realizable AEs and otherwise, unrealizable AEs.
Here we calculate the CSR for each example based on all its features rather than only the added adversarial features because, in practice, it is not known which of all features are added by an attack. Figure~\ref{fig:acc_differentiation} shows the results with varied CSR thresholds and $\epsilon$ bounds.
As can be seen, the CSR threshold should be high enough to differentiate realizable AEs from unrealizable AEs.
However, the CSR should not be set too high because our learned domain constraints may not perfectly represent the actual domain constraints.
In addition, the differentiation is easier when the $\epsilon$ bounds are high (e.g., $\epsilon=20$ and $\epsilon=30$).
Note that a lower $\epsilon$ requires a higher threshold for optimal accuracy. This is because a lower $\epsilon$ implies fewer perturbed features, which means more features remain the same as the original features, leading to a higher CSR baseline.

\begin{table}[!b]
\caption{TPR and FPR of the proposed AE detection method against various evasion attacks.}
\begin{center}
\begin{tabular}{l|cc}
\toprule
\textbf{Attack}
&\textbf{TPR}
&\textbf{FPR}\\
\midrule
GenDroid & 97.5 & 6.9\\
ShadowDroid & 95.2 & 4.3\\
Grosse Attack & 75.4 & 6.9\\
\bottomrule
\end{tabular}
\label{table:ae_detection}
\end{center}
\end{table}

Finally, we assess our defense, introduced in \S\ref{subsection:ae_detection}, to show the impact of our learned domain constraints on identifying AEs generated by three evasion attacks: GenDroid, ShadowDroid, and Grosse Attack, whose resulting AEs may not be realizable. Specifically, we prepare an evaluation set comprising $5K$ benign samples and $1K$ malware samples randomly selected from the test set. Depending on the attack under investigation, AEs of malware samples generated by GenDroid, ShadowDroid, or Grosse Attack targeting DREBIN-DNN are also included in the evaluation set. All attacks are constrained by an $L_1$ norm bound of $\epsilon=30$ for AE generation. Furthermore, our preliminary evaluation indicates that setting the CSR threshold at 92\% can result in high detection rates of AEs. Table~\ref{table:ae_detection} demonstrates the capability of our proposed AE detection method to effectively identify AEs. It is worth noting that utilizing a smaller CSR threshold can decrease the occurrences of false detections, albeit at the expense of reducing AE detection rates.

\begin{tcolorbox}[sharp corners,colframe=black,boxrule=0.2mm]
\textbf{RQ1.} 
Can our learned domain constraints be applied to counter evasion attacks that generate AEs without ensuring their feasibility?

\textbf{Yes, because our learned domain constraints are capable of distinguishing between realizable AEs and unrealizable AEs.}
\end{tcolorbox}

\begin{table}[!b]
\caption{The average model utility of different malware detectors based on five trials. DREBIN-DNN, DroidAPIMiner-DNN, and RAMDA-DNN operate on binary feature spaces, whereas R-PackDroid-DNN operates on a non-binary feature space.}
\begin{center}
\begin{tabular}{l|lccc}
\toprule
\textbf{Detector}
&\textbf{Defense}
&{\textbf{TPR}}
&{\textbf{FPR}}
&{\textbf{\shortstack{Clean Acc}}}\\
\midrule
\multirow{4}{*}{\shortstack{DREBIN-DNN}} &
 W/O Defense & 81.0\% & 0.4\% & 96.4\%  \\
& AT-Unrealizable-AEs & 79.6\% &  0.4\% & 96.2\% \\
& AT-Non-Uniform-Perturbations & 81.2\% & 0.4\% & 96.5\% \\
& AT-Realizable-AEs (ours) & 81.0\% & 0.4\% & 96.3\% \\
\midrule
\multirow{4}{*}{\shortstack{DroidAPIMiner-DNN}} &
 W/O Defense & 77.2\%  & 1.0\% & 95.4\%  \\
& AT-Unrealizable-AEs & 76.1\% & 1.0\%  &  95.2\%\\
& AT-Non-Uniform-Perturbations & 74.5\% & 1.0\% & 94.9\% \\
& AT-Realizable-AEs (ours) & 75.1\% & 1.0\% & 95.0\%\\
\midrule
\multirow{4}{*}{\shortstack{RAMDA-DNN}} &
 W/O Defense & 86.1\%  & 0.9\% & 96.9\%  \\
& AT-Unrealizable-AEs & 84.9\% & 0.8\% &  96.8\%\\
& AT-Non-Uniform-Perturbations & 85.3\% & 0.9\% & 96.7\% \\
& AT-Realizable-AEs (ours) & 84.0\% & 0.9\% & 96.6\%\\
\midrule
\multirow{4}{*}{\shortstack{R-PackDroid-DNN}} &
 W/O Defense & 80.4\%  & 0.9\% & 95.9\%  \\
& AT-Unrealizable-AEs & 77.7\% & 1.1\% &  95.4\%\\
& AT-Non-Uniform-Perturbations & 77.4\% & 1.0\% & 95.4\% \\
& AT-Realizable-AEs (ours) & 78.0\% & 1.0\% & 95.5\%\\
\bottomrule

\end{tabular}
\label{table:clean-acc}
\end{center}
\end{table}

\begin{figure}[t]
    \centering
    \begin{subfigure}[b]{0.35\textwidth} 
        \caption{DREBIN-DNN}
        \centering
        \includegraphics[width=\textwidth]{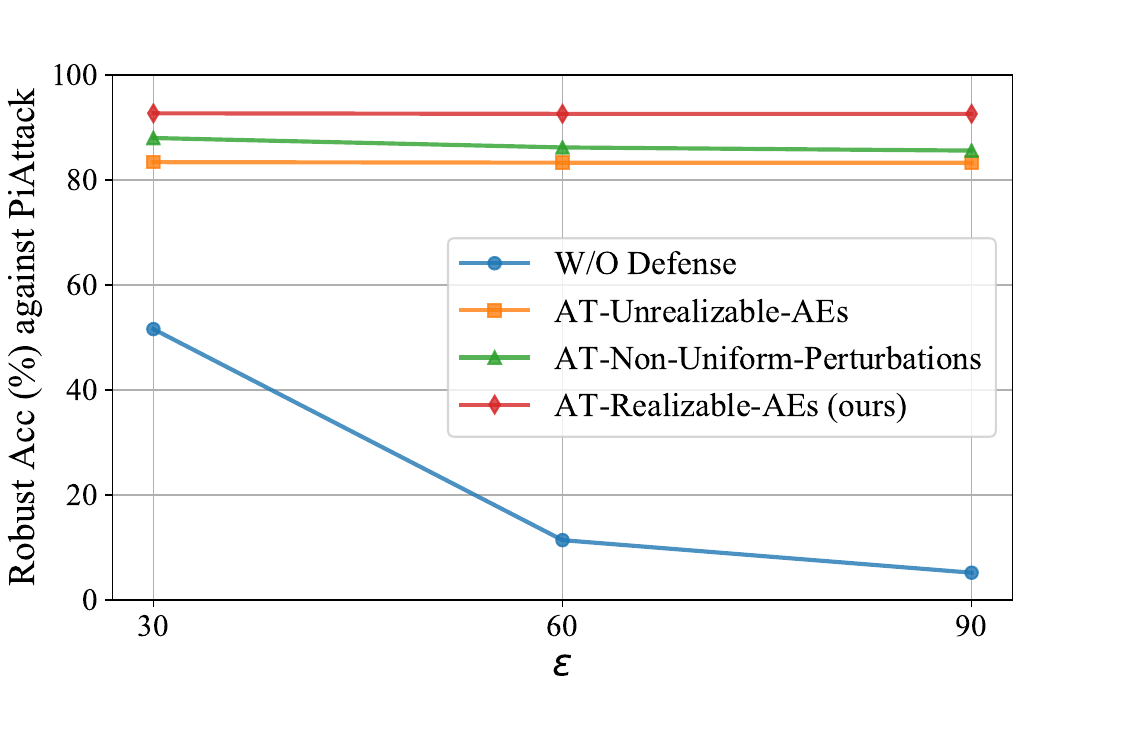}        
        \label{fig:image1}
    \end{subfigure}
    \hspace{-0.5em} 
    \begin{subfigure}[b]{0.35\textwidth} 
        \caption{DREBIN-DNN}
        \centering
        \includegraphics[width=\textwidth]{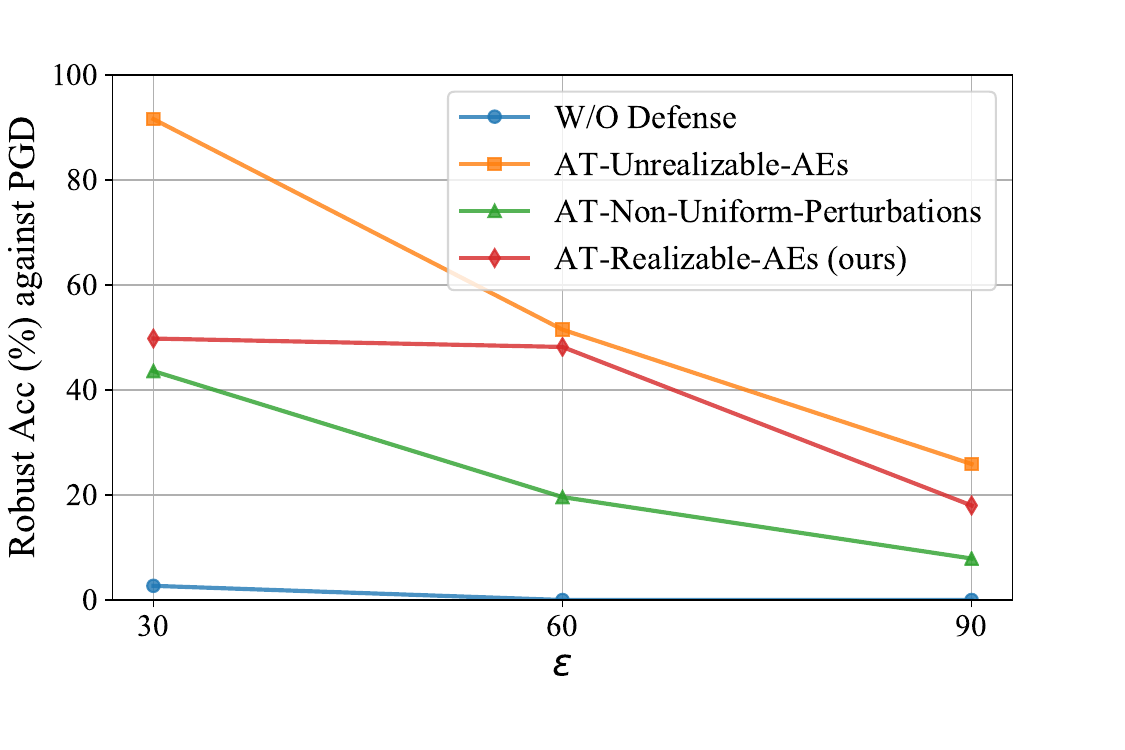}        
        \label{fig:image2}
    \end{subfigure}

    \vspace{-2em} 

    \begin{subfigure}[b]{0.35\textwidth} 
        \caption{DroidAPIMiner-DNN}
        \centering
        \includegraphics[width=\textwidth]{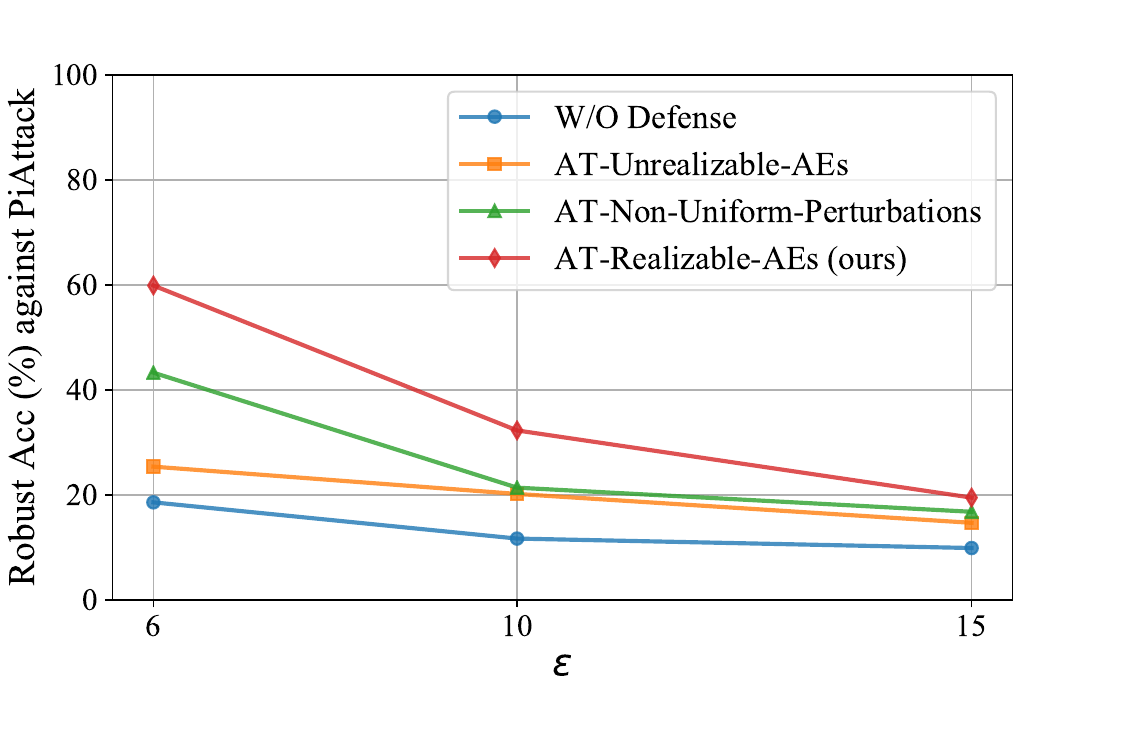}        
        \label{fig:image5}
    \end{subfigure}
    \hspace{-0.5em} 
    \begin{subfigure}[b]{0.35\textwidth} 
        \caption{DroidAPIMiner-DNN}
        \centering
        \includegraphics[width=\textwidth]{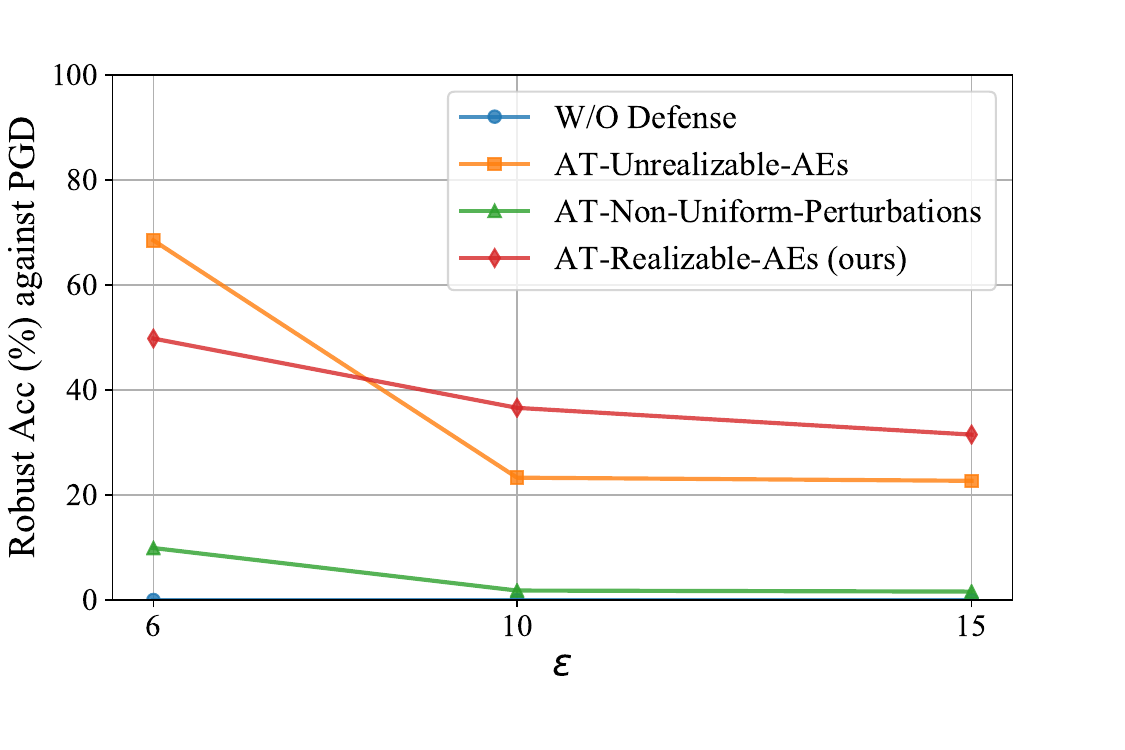}        
        \label{fig:image6}
    \end{subfigure}
    
    \vspace{-2em} 

    \begin{subfigure}[b]{0.35\textwidth} 
        \caption{RAMDA-DNN}
        \centering
        \includegraphics[width=\textwidth]{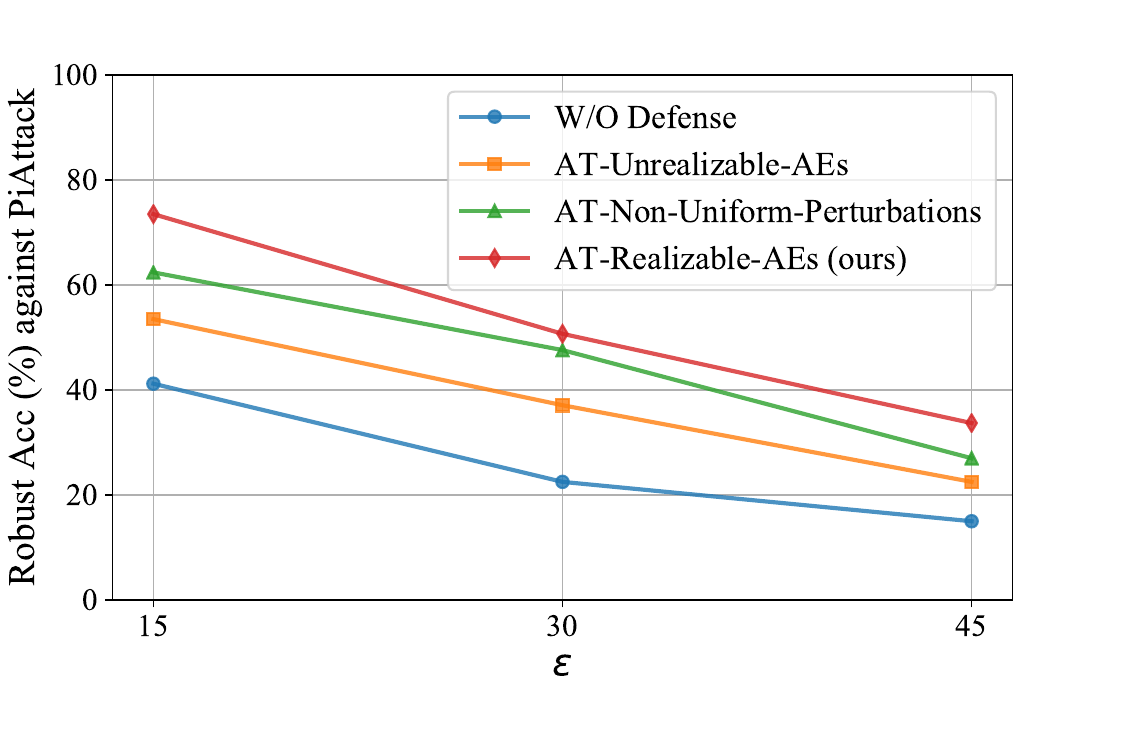}        
        \label{fig:image3}
    \end{subfigure}
    \hspace{-0.5em} 
    \begin{subfigure}[b]{0.35\textwidth} 
        \caption{RAMDA-DNN}
        \centering
        \includegraphics[width=\textwidth]{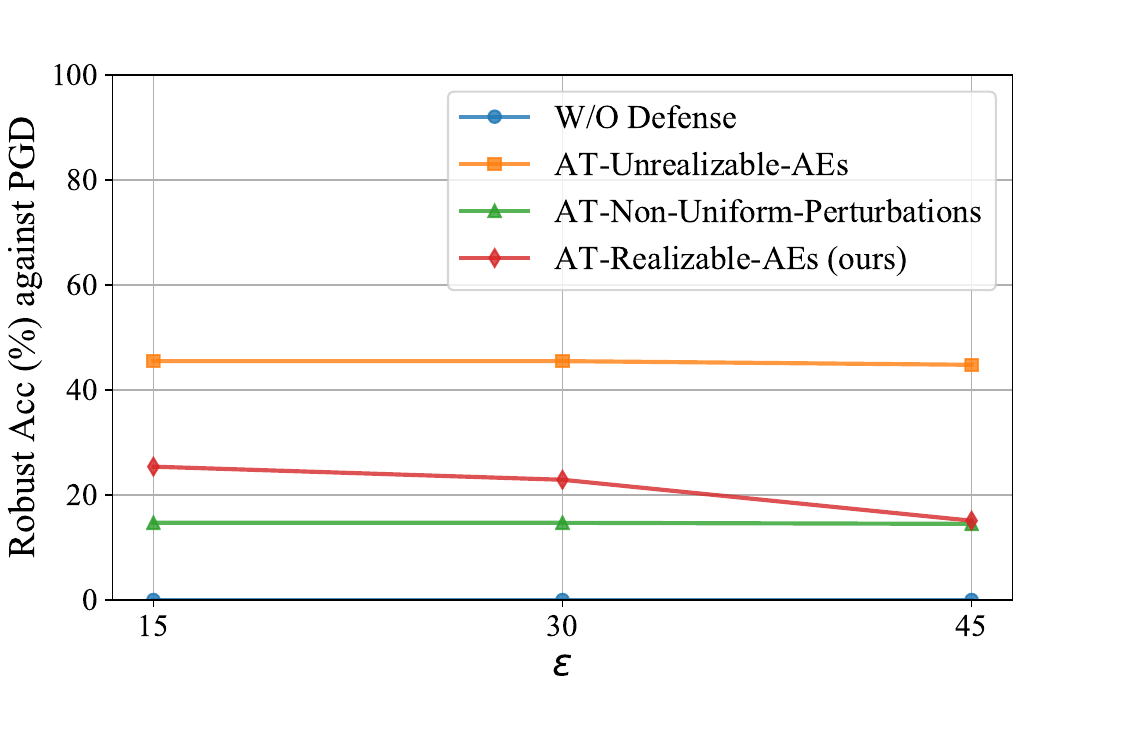}        
        \label{fig:image4}
    \end{subfigure}
    
    \vspace{-2em} 

    \begin{subfigure}[b]{0.35\textwidth} 
        \caption{R-PackDroid-DNN}
        \centering
        \includegraphics[width=\textwidth]{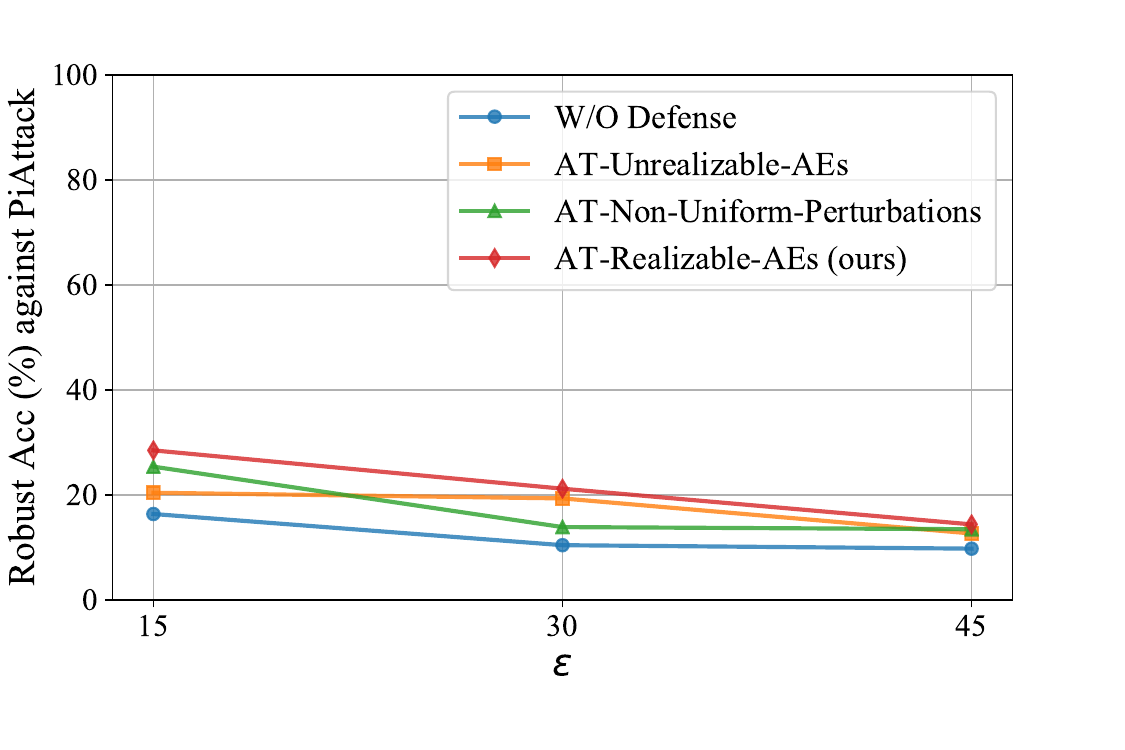}       
        \label{fig:image7}
    \end{subfigure}
    \hspace{-0.5em} 
    \begin{subfigure}[b]{0.35\textwidth} 
        \caption{R-PackDroid-DNN}
        \centering
        \includegraphics[width=\textwidth]{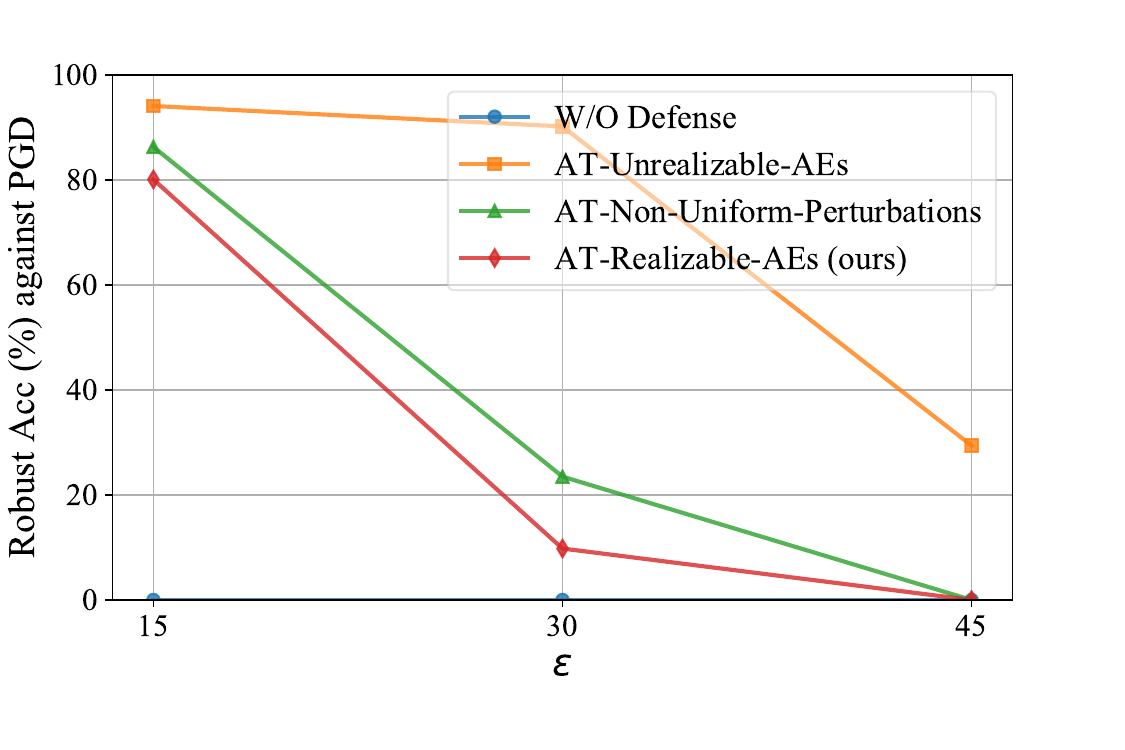}        
        \label{fig:image8}
    \end{subfigure}
    \vspace{-2.5em} 
    \caption{The adversarial robustness of different detectors against both an unrealistic attack (PGD) and a realistic attack (PiAttack). Results are averaged over five trials. DREBIN-DNN, DroidAPIMiner-DNN, and RAMDA-DNN operate on binary feature spaces, whereas R-PackDroid-DNN operates on a non-binary feature space.}
    \label{fig:robustness-models-with-adversarial-retraining}
\end{figure}

\subsection{Evaluating Our Defense}
\label{section:AT-evaluation}
In order to answer \textbf{\hyperlink{RQ2}{RQ2}} 
stated in \S\ref{section:simulation-results}, this section empirically evaluates our defense, which is based on AT with realizable AEs generated by considering feature-space domain constraints, as introduced in \S\ref{section:adversarial_training}.
Specifically, DREBIN-DNN, DroidAPIMiner-DNN, RAMDA-DNN, and R-PackDroid-DNN are trained with different strategies: standard training indicating w/o defense, AT with unrealizable AEs, and our AT with realizable AEs. We also consider a state-of-the-art AT strategy~\cite{b38} that relies on non-uniform perturbations, denoted as AT-Non-Uniform-Perturbations in our experiment.
We refer the reader to Appendix~\ref{apendix:non-uniform-perturbations} for further details about the AT-Non-Uniform-Perturbations approach. For all three AT approaches, PGD~\cite{b21} is adopted to generate AEs in every training epoch. Specifically, half of the training malware samples are used to generate AEs and the rest remain unmodified. To generate realizable AEs in the R-PackDroid feature space, we use random forest regression, which can capture the complex relationships, to build the regression models mentioned in Algorithm~\ref{algorithm:generate_ae}. Our preliminary evaluation shows that PiAttack requires adding an average of 30 new features to achieve a successful AE in attacking DREBIN-DNN, 6 new features for DroidAPIMiner-DNN, and 15 new features for both RAMDA-DNN and R-PackDroid-DNN. Therefore, we adopt $L_1$ norm bound with $\epsilon=30$ for DREBIN-DNN, $\epsilon=6$ for DroidAPIMiner-DNN, and $\epsilon=15$ for both RAMDA-DNN and R-PackDroid-DNN. It is worth emphasizing that the perturbation bounds for all AT approaches are the same. Moreover, since AT uses minibatch Stochastic Gradient Descent to train DNN models, there is inherent randomness in the selection of samples, particularly malware samples, in each batch. To account for this randomness and reduce bias, we perform the experiment across five trials. This involves repeating the experiments five times and reporting the average results for both the model's performance on clean data and its adversarial robustness.

Table~\ref{table:clean-acc} shows the performance of different detectors on clean samples and Figure~\ref{fig:robustness-models-with-adversarial-retraining} reports their robustness.
For robustness, we test both the unrealistic attack, PGD, and the realistic attack, PiAttack, and vary the $\epsilon$ values for both attacks. Considering large perturbation bounds beyond norm-bounded perturbations can provide insights into the detector's performance against realistic attacks, which normally may succeed with large perturbations.
We make sure that the AEs are generated from the malware samples that were correctly detected by malware detectors. As can be seen, in general, different defenses yield similar clean performance (Table~\ref{table:clean-acc}) but different robustness (Figure~\ref{fig:robustness-models-with-adversarial-retraining}).
For DREBIN-DNN, our defense achieves the best robustness with an accuracy of 92.7\%, thus surpassing the performance of AT-Unrealizable AEs and AT-Non-Uniform-Perturbations for $\epsilon=30$. Specifically, for larger values of $\epsilon$, the improvement of our proposed approach over other robust detectors, especially over AT-Non-Uniform-Perturbations slightly increases. For DroidAPIMiner-DNN, RAMDA-DNN, and R-PackDroid-DNN, our defense is still the best, e.g., 59.9\% vs. 43.3\% for AT-Non-Uniform-Perturbations at $\epsilon = 6$ when the target detector is DroidAPIMiner, 73.5\% vs. 62.4\% for AT-Non-Uniform-Perturbations at $\epsilon = 15$ when the target detector is RAMDA-DNN, and 28.5\% vs. 25.4\% for AT-Non-Uniform-Perturbations at $\epsilon = 15$ when the target detector is R-PackDroid-DNN. Figure~\ref{fig:robustness-models-with-adversarial-retraining} also demonstrates that evaluating detectors against unrealistic attacks may not accurately reflect the actual robustness against realistic attacks. As an example, with DREBIN-DNN as the target detector at $\epsilon=30$, it would slightly overestimate the robustness of AT-Unrealizable AEs but largely underestimate the robustness of AT-Non-Uniform-Perturbations and our AT-Realizable AEs.
\begin{tcolorbox}[sharp corners,colframe=black,boxrule=0.2mm]
\textbf{RQ2}. Can our learned domain constraints help AT enhance the robustness of the detectors against realistic evasion attacks?

\textbf{Yes, incorporating our learned domain constraints provides AT with feature-space realizable AEs, which are more effective than AT with feature-space unrealizable AEs}.
\end{tcolorbox}

\subsection{Evaluating Our OPF-based Method}
\label{sec:evaluate_OPF}
In this experiment, we aim to answer \textbf{\hyperlink{RQ3}{RQ3}} stated in \S\ref{section:simulation-results} by validating the ability of OPF to extract meaningful feature dependencies. We compare the proposed OPF-based method with a straightforward baseline method that is based on threshold clustering (TC). For a specific feature $f_a$, TC exclusively chooses the dependent features from its top-N dependent features.
As a sanity check, we also report the results for TC with bottom-N dependent features, where the least dependent features are used. This experiment specifically compares the adversarial robustness of DREBIN-SVM retrained based on the defense introduced in \S\ref{section:adversarial-retraining} by changing 500 malware samples of the training data set to AEs when OPF- or TC-based approach is used to make them realizable. 
Note that for re-training DREBIN-SVM with AEs, we use PK-Feature under our learned domain constraints.
Here we measure the robustness of different augmented DREBIN-SVMs against transferable problem-space realizable AEs generated by PiAttack using the original DREBIN-SVM as the surrogate detector.
This is because generating problem-space AEs directly on different augmented DREBIN-SVMs in a white-box setting is time-consuming. Moreover, we make sure that the AEs are generated from the malware samples correctly detected by all augmented DREBIN-SVMs, and the results are calculated on successful AEs for the original DREBIN-SVM.

Table~\ref{table:performance-of-augmented-DREBIN-FeatureIdeal-models} shows that the clean accuracy for all three detectors is comparable.
For robustness, although the TC method confirms the effectiveness of considering feature correlations in identifying feature dependencies that enhance adversarial robustness, our proposed OPF method surpasses it ($82.9\%$ vs. $69.4\%$ under Top-80).
In contrast, as expected, using bottom-N features leads to significantly worse results.
It is also worth mentioning that the TC method requires careful tuning (e.g., via linear search) of the hyperparameter $N$ to achieve the best performance but our OPF approach does not. 

\begin{tcolorbox}[sharp corners,colframe=black,boxrule=0.2mm]
\textbf{RQ3.} Can our OPF-based method effectively extract meaningful feature dependencies?

\textbf{Yes, OPF notably outperforms the baseline that simply selects top-N dependent features.}
\end{tcolorbox}

\begin{table}[!t]
\caption{The model utility and robustness of DREBIN-SVM augmented by OPF-based method vs. TC-based method.} 
\begin{center}
\begin{tabular}{lrccc}
\toprule
{\textbf{Detector}}&
{\textbf{N}}&
{\textbf{\shortstack{Clean Acc~(\%)}}}&{\textbf{\shortstack{Robust Acc~(\%)}}}\\
\midrule
\textbf{OPF-DREBIN-SVM} & N/A & 96.6\% & 82.9\%\\
\midrule
\multirow{5}{*}{\textbf{\shortstack{TC-DREBIN-SVM\\(Top-N)}}} & 20 & 96.6\% & 45.7\% \\
 & 40 & 96.7\% & 64.6\% \\
 & 60 & 96.6\% & 65.2\% \\
 & 80  & 96.8\% & 69.4\% \\
 & 100 & 96.7\% & 57.1\% \\
 \midrule
 \multirow{5}{*}{\textbf{\shortstack{TC-DREBIN-SVM\\(Bottom-N)}}} & 20 & 96.7\% & 12.9\% \\
 & 40 & 96.7\% & 13.8\%\\
 & 60 & 96.7\% & 24.0\%\\
 & 80 & 96.6\% & 13.6\%\\
 & 100 & 96.6\% & 11.6\%\\
\bottomrule
\end{tabular}
\label{table:performance-of-augmented-DREBIN-FeatureIdeal-models}
\end{center}
\end{table}

\subsection{Feature-Space Realizable AEs vs. Problem-Space Realizable AEs}
\label{sec:feature_vs_problem_space_comparison}
To explore \textbf{\hyperlink{RQ4}{RQ4}} stated in \S\ref{section:simulation-results}, we assess how using different types of realizable AEs impacts the training speed and generalizability of ML-based malware detection. Note that when investigating generalizability, we aim to understand how well a detector, enhanced with realizable AEs, can resist problem-space attacks that generate realizable AEs distinct from those used during model training. In other words, our objective is to evaluate the robustness of a detector hardened with certain transformations against attacks utilizing new transformations. To accomplish this, we retain some problem-space transformations exclusively for the purpose of attacking, while others are available for use during the training phase. Specifically, we randomly select a subset of collected problem-space transformations. We then employ PiAttack with this subset of transformations to convert 500 malware samples in the training set into problem-space realizable AEs. Subsequently, we expand our training set by incorporating these problem-space realizable AEs and proceed to retrain DREBIN-SVM. We follow a similar process to retrain DREBIN-SVM with our feature-space realizable AEs, which are generated by PK-Feature operating under our learned domain constraints.
From Table~\ref{table:speed_generalizability}, we can observe how various retraining strategies perform in terms of training speed and the level of robustness they offer. PiAttack directly targets DREBIN-SVM hardened with different defenses to generate problem-space realizable AEs from a set of 500 malware apps. In Table~\ref{table:speed_generalizability}, NoF indicates the number of AEs, R\_Time denotes the retraining time, and C\_Acc represents the clean accuracy. Moreover, R\_Acc1 and R\_Acc2 denote robust accuracy against realizable AEs generated by a subset of problem-space transformations, which are also employed during retraining, and against all problem-space transformations, respectively. Table~\ref{table:speed_generalizability} demonstrates the significant increase in computational complexity when retraining DREBIN-SVM with problem-space realizable AEs compared to the scenario where we utilize feature-space AEs for retraining.

Furthermore, it's evident that AT-PiAttack primarily enhances robustness against realizable AEs that resemble those used during retraining. However, this robustness significantly diminishes when faced with realizable AEs that might differ from those employed in adversarial retraining. On the other hand, employing feature-space realizable AEs during retraining aids in maintaining the detector's robustness. In fact, utilizing feature-space domain constraints enables us to rapidly generate a greater number of realizable AE variants from each malware sample, thereby uncovering more vulnerable regions during the retraining process. As depicted in Table~\ref{table:speed_generalizability}, our AT-PK-Feature defense, which retrains DREBIN-SVM with $\approx 10K$ feature-space realizable AEs, exhibits strong robust accuracy against PiAttack when it can utilize all available transformations. It is worth noting that generating a large number of additional (problem-space) Realizable AEs is not practical for AT-PiAttack due to its high computational cost.

\begin{table}[!t]
\caption{The performance of different defenses used in DREBIN-SVM in terms of re-training time (s), clean and robust accuracy (\%). 
}
\begin{center}
\begin{tabular}{lrrrrr}

\toprule
{\textbf{Defense}}
&{\textbf{\shortstack{NoA}}}
&{\textbf{\shortstack{R\_Time}}}
&{\textbf{\shortstack{C\_Acc}}}&{\textbf{\shortstack{R\_Acc1}}}&{\textbf{\shortstack{R\_Acc2}}}\\
\midrule
AT-PiAttack & 500  & 110,543s  & 96.6\% & 95.6\% & 5.0\%\\
\midrule
AT-PK-Feature (ours) & \multirow{2}{*}{500} & 224s & 96.7\% & 94.2\% & 20.8\%\\
AT-PK-Feature-un & & 187s & 96.7\% & 83.2\% & 14.8\%\\
\midrule
AT-PK-Feature (ours) & \multirow{2}{*}{$\approx 10K$} & 1,559s & 96.7\% & 100.0\% & 82.9\%\\
AT-PK-Feature-un && 952s & 96.6\% & 85.4\% & 37.4\%\\
\bottomrule
\end{tabular}
\label{table:speed_generalizability}
\end{center}
\end{table}

\begin{tcolorbox}[sharp corners,colframe=black,boxrule=0.2mm]
\textbf{RQ4.} Can our feature-space realizable AEs outperform the conventional problem-space realizable AEs?

\textbf{Yes, feature-space realizable AEs yield higher AT training efficiency and generalizability of the detector.}
\end{tcolorbox}

\subsection{Discussion}
In order to improve the adversarial robustness of malware detection, it is necessary to provide a realistic view of the vulnerabilities of ML-based detectors to realizable AEs, which are generated under domain constraints of malware apps~\cite{b1}. Our experimental results derived from various experiments have demonstrated the general effectiveness of our feature-space solution in hardening Android malware detection against evasion attacks. Note that our experiments are designed to empirically assess the efficacy of different aspects of our feature-space solution. The proposed solution is structured around a three-fold approach, which includes (1) establishing feature-space domain constraints by analyzing various aspects of Android malware properties, (2) learning feature-space domain constraints from large-scale data, and (3) applying the learned feature-space domain constraints to fortifying the robustness of AMD against evasion attacks. Specifically, our analysis in \S\ref{section:evaluate-realizable-feature-space-perturbations-in-adversarial-retraining} and \S\ref{sec:evaluate_OPF} validates the performance of our solution for the first and second folds. Moreover, our assessment in \S\ref{section:evaluate-realizable-feature-space-perturbations-in-adversarial-retraining}, \S\ref{section:AT-evaluation}, and \S\ref{sec:feature_vs_problem_space_comparison} provides compelling evidence in support of the third fold. Here we further discuss the advantages of our solution from three main aspects: practicality, generalizability, and detectability.

\begin{table}[!b]
\caption{Computational time required to generate AEs using various attacks.}
\label{table:computaional_time}  
\begin{center}
\begin{tabular}{llrr}
\toprule
\textbf{Attack} & \textbf{Attack Surface} & \textbf{\shortstack{Avg. Time for\\ Generating AE}} & \textbf{\shortstack{Avg. Time \\ per Modification}} \\
\midrule
\textbf{PiAttack} & Problem space & 221.09s & 44.2s \\
\midrule
\multirow{2}{*}{\shortstack[l]{\textbf{Our approach}}} & Binary feature space & 0.44s & 0.09s \\
 & Non-binary feature space& 0.81s & 0.16s \\
\bottomrule
\end{tabular}
\end{center}
\end{table}

\noindent\textbf{Practicality.} Our findings in \S\ref{section:AT-evaluation} and \S\ref{sec:feature_vs_problem_space_comparison}, indicate that generating feature-space AEs to incorporate in adversarial hardening
can be a promising alternative to generating time-consuming problem-space AEs, particularly when they successfully satisfy domain constraints. Generally, PiAttak requires an average of 5 problem-space transformations to convert a malware app into a problem-space AE. Each transformation involves several steps: (i) choosing a gadget (i.e., a slice of an app’s bytecode) that indicates a problem-space transformation and loading it from disk, (ii) injecting the gadget into the malware app using Soot~\cite{soot}, (iii) performing static analysis using Apktool~\cite{apktool} (i.e., a reverse engineering tool), and (iv) constructing the feature representation based on the extracted features in the feature space of the malware classifier. Although all of these steps involve time and computational overhead, gadget injection and static analysis are particularly time-consuming compared to other processes, each averaging around 20 seconds. In the PiAttak, memory overhead is also an important consideration alongside computational time. Each step involved in transforming a malware app into an adversarial app contributes to the overall memory usage as follows:

\begin{itemize}
    \item Load and inject the gadget: Loading a gadget from the disk and injecting it into the app requires additional memory. The size of the gadget and its integration with the app's existing codebase can impact memory consumption. For instance, loading a $1~MB$ gadget into memory might require slightly more than $1~MB$ due to overheads associated with loading and managing data structures.
    \item Static analysis: performing static analysis with tools like Apktool involves parsing and analyzing the app’s APK file. This process can be memory-intensive, as the tool needs to maintain a representation of the app's structure and resources in memory during analysis. For instance, for a $5~MB$ app, the static analysis tool could require around ~$15~MB$ ($5~MB$ app size × 3 for analysis overhead).
\end{itemize}

Generating AEs in the feature space eliminates the need to conduct steps (i) to (iv) for each modification, thereby significantly accelerating the speed of AT. As shown in Table~\ref{table:speed_generalizability}, the retraining time for DREBIN-SVM is 110,543 seconds when using 500 problem-space realizable AEs for hardening. In contrast, the retraining time is only 224 seconds with 500 feature-space realizable AEs. Specifically, our empirical analysis in Table~\ref{table:computaional_time} demonstrates that creating a feature-space realizable AE in a binary features space takes only 0.44 seconds, whereas generating a problem-space realizable AE takes 221.09 seconds. This remarkable improvement is attributed to the fact that our approach eliminates the time-consuming process of generating problem-space AEs based on transformations. Our analysis reveals that generating feature-space AEs in a non-binary feature space takes around 0.81 seconds, which is still significantly quicker than the 221.09 seconds needed to generate problem-space AEs. Specifically, constructing the regression model, the most time-consuming phase in generating feature-space AEs, requires about $0.03$ second with random forest regression using $10$ decision trees on a $120K$ sample training set. This time is negligible when compared to the significant computational effort needed for problem-space AEs. Additionally, building a regression model is not always required, as a pre-constructed model for certain independent and dependent features may already be available, as outlined in Algorithm~\ref{algorithm:generate_ae}. 
In summary, the significant improvement in generating AEs within the feature space during AT indicates the potential use of feature-space realizable AE in real-world scenarios where defenders need to harden large Android malware detectors with AT.

It noted that in the preprocessing stage, we need to measure the correlation between each pair of features in the dataset. For low-dimensional datasets like DroidAPIMiner, this process takes only a few seconds. However, for very high-dimensional datasets like DREBIN, it can take over 10 hours on our Debian Linux workstation equipped with an Intel (R) Core (TM) i7-4770K CPU at 3.50 GHz and 32 GB of RAM. Despite the longer preprocessing time for high-dimensional datasets, this process is still significantly more efficient than the processing required for problem-space attacks. For example, collecting around 500 problem-space transformations, including program slicing of gadgets from benign apps, can take more than a week.


\noindent\textbf{Generalizability.} As examined in \S\ref{sec:feature_vs_problem_space_comparison}, our feature-space realizable AEs exhibit high generalizability as there are no limitations on the generation of realizable AEs in the feature space. On the other hand, generating problem-space realizable AEs has specific limitations because they rely on limited sets of problem-space transformations~\cite{b29,b30,b81}. This means it is possible that a realistic attack based on a new set of transformations can bypass the detector that is adversarially trained on those limited sets~\cite {b1}. In contrast, feature-space realizable AEs can potentially generate more diverse realizable AEs when they take into account the domain constraints in the feature space. For instance, our empirical analysis shows that for each problem-space realizable AE generated by PiAttack, we can generate about 20 different variants of feature-space realizable AEs using PK-Feature attack under our learned domain constraints.

\noindent\textbf{Detectability.} The results outlined in \S\ref{section:evaluate-realizable-feature-space-perturbations-in-adversarial-retraining} illustrate that our learned domain constraints are able to function as a preprocessing method in detecting AEs prior to engaging an ML-based malware detector. Specifically, this non-ML defense entails a thorough analysis of the feature representations of apps to uncover suspicious apps that are not practically feasible by determining their violations from the predefined domain constraints.
\section{Limitations and Future Work}
\label{section:limitations-and-future-work}
While we have demonstrated the effectiveness of our proposed defensive approach through extensive experiments, there are some limitations that need further investigation. 

First, the proposed technique works for a subset of detectors that rely on expressive, domain-specific features (e.g., API calls and permissions), which capture meaningful dependencies in Android apps. Within this subset, we have specifically explored both binary and non-binary features. Future work should explore other detectors that rely on more basic features, such as byte sequences~~\cite{daoudi2021dexray} or opcode analysis~\cite{jerome2014using}.

Second, like any data-driven technique, our approach may be potentially biased toward the specific training data because data might be inaccurate and incomplete. For instance, as shown in Table~\ref{table:distribution_drift}, the purely data-driven approaches might intensify the \textit{concept drift} issue~\cite{b84}, which is a common challenge in the ML context. 
Here, we empirically illustrate this limitation by testing different DREBIN-DNNs introduced in \S\ref{section:AT-evaluation} on $15K$ ($12K$ benign and $3K$ malware) newly collected Android apps from AndroZoo~\cite{b67}. These new samples have been released between 2020 and 2022. As can be seen from Table~\ref{table:distribution_drift}, all of the newly measured clean accuracies are reduced compared to the old results, indicating the existence of concept drift. While concept drift can also affect adversarial robustness, Table~\ref{table:distribution_drift} demonstrates that the proposed approach still achieves better robust accuracy than other detectors, confirming its effectiveness in providing adversarial robustness against newer adversarial malware apps.

Our study shows that feature correlations are theoretically adequate for identifying meaningful dependencies that reflect domain constraints in the feature space, but their success depends on the quality of the training data. Feature correlations can perform well when the utilized dataset accurately represents the true data distribution. However, as indicated in Table~\ref{table:distribution_drift}, correlations may vary if the dataset does not reflect the true data distribution. To address this concept drift, techniques like continuous learning can help maintain accuracy by adapting to current distributions of malware and benign apps.
Incorporating domain knowledge could also address this limitation.
However, a key question is how the knowledge of domain experts can be incorporated to specify domain constraints in the feature space. One can assume that domain knowledge can make the search for feature dependencies more precise; therefore, an interesting avenue for future study is to incorporate domain knowledge to complement the data-driven approaches in finding meaningful feature dependencies.

\begin{table}[!t]
    \centering
    \caption{The clean and robust accuracy of different defenses used in DREBIN-DNN 
    on old/new test samples. The new samples have been released between 2020 and 2022. The hardened models are strengthened using PGD with 
    $\epsilon=30.$}    
    \begin{tabular}{l|cc|cc}
    \toprule
    \multirow{2}{*}{\textbf{ Defense}} & \multicolumn{2}{c}{\textbf{ Clean Acc}} & \multicolumn{2}{c}{\textbf{ Robust Acc}} \\
    \cline{2-5}     
      & \textbf{\shortstack{Old Test Set}}     
      & \textbf{ \shortstack{New Test Set}} 
      & \textbf{\shortstack{Old Test Set}}     
      & \textbf{\shortstack{New Test Set}}\\
    
    \midrule
    W/O Defense &  96.4\% &  89.1\% & 51.6\% & 32.0\%\\
    AT-Unrealizable-AEs &  96.2\% & 88.1\% & 83.4\% & 73.2\%\\
    AT-Non-Uniform-Perturbations & 96.5\% &  86.7\% & 88.0\% & 77.4\%\\
    AT-Realizable-AEs (ours) & 96.3\% &  85.2\% & 92.7\% & 80.2\%\\
    \bottomrule
    \end{tabular}    
    \label{table:distribution_drift}
    \end{table}
\section{Conclusion}
\label{section:conclusion}
In this paper, we propose a new approach to facilitate uncovering vulnerable regions within ML models employed in AMD, consequently enhancing the capability of defense mechanisms against evasion attacks, particularly realistic attacks.
Specifically, we present a new interpretation of domain constraints in the feature space by extracting meaningful feature dependencies.
To this end, we not only consider statistical correlations but also adopt OPF to extract such dependencies and apply them either in AE detection to identify feasible AEs or in generating feature-space RealAEs during AT to improve the robustness of detectors against RealAEs.
The empirical results show the general effectiveness of our new approach in strengthening the robustness of AMD. In particular, our assessment underscores the superior efficiency and generalizability of our defense in comparison to problem-space RealAEs when it comes to adversarial hardening. Additionally, our extracted feature dependencies have proven effective in distinguishing between feasible and unfeasible ones such as unRealAEs, thereby demonstrating the significant potential for use as a reliable criterion for defenses that work based on AE detection.

\begin{acks}
Zhengyu Zhao was supported by the National Natural Science Foundation of China under the grant 62406240.
Zhuoran Liu was (in part) supported by the Dutch Research Council (NWO) through the PROACT project (NWA.1215.18.014), TTW PREDATOR project 19782, and the CiCS project of the research program Gravitation under the grant 024.006.037. Veelasha Moonsamy was supported by the Deutsche Forschungsgemeinschaft (DFG, German Research Foundation) under Germany’s Excellence Strategy - EXC 2092 CASA - 390781972. 
\end{acks}

\bibliographystyle{unsrt}
\bibliography{manuscript}

\appendix
\section{Evaluating the Efficacy of Learned Domain Constraints with Sparse-RS}
\label{apendix:sparse_rs}
To further assess the effectiveness of our learned domain constraints in enhancing the adversarial robustness of malware detectors, we utilize another adversarial attack known as Sparse-RS~\cite{b62}, which is inherently suitable for generating AEs in discrete domains like malware detection. Sparse-RS operates as a sparse adversarial attack within black-box settings, employing a heuristic search strategy—specifically, a randomized search that is effective in both discrete and continuous feature spaces. This approach not only identifies AEs with minimal changes to input features but also achieves query efficiency by reducing interactions with the target detectors.

For our evaluation, we set the initial decay factor $\alpha_{init}=1.6$, sparsity level $k=180$ as per~\cite{b62}, and a query budget $Q=100$. Table~\ref{table:sparse_rs} demonstrates that incorporating our learned domain constraints in AT to make AEs generated by Sparse-RS realizable significantly enhances adversarial robustness compared to using typical AEs generated by Sparse-RS. Moreover, our approach interestingly maintains performance on clean data better than another defense method.

\begin{table}[!b]
\caption{The clean performance and adversarial robustness of DREBIN-DNN detectors hardened through adversarial training (AT) using Sparse-RS, with and without incorporating our domain constraints at $\epsilon=30$. The robust accuracy is measured against PiAttack with various attack bounds.} 
\begin{center}
\begin{tabular}{lcccccc}
\toprule
{\textbf{Defense}}&
{\textbf{TPR}}&
{\textbf{FPR}}&
{\textbf{\shortstack{Clean Acc}}}& \multicolumn{3}{c} {\textbf{Robust Acc}}\\
\cline{5-7} &
&&& $\epsilon=30$ & $\epsilon = 60$ & $\epsilon = 90$\\
\midrule
{\textbf{AT-Unrealizable-AEs}} & 73.9\% & 0.3\% & 95.3\% & 52.0\% & 46.2\%& 39.3\%\\
 {\textbf{AT-Realizable-AEs (ours)}} & 80.8\% & 0.4\% & 96.3\% & 59.9\% & 50.9\% & 48.1\%\\
\bottomrule
\end{tabular}
\label{table:sparse_rs}
\end{center}
\end{table}

\section{PiAttack}
\label{apendix:PiAttack}
PiAttack~\cite{b17} is an adversarial attack that operates in the problem space and transforms Android apps into adversarial ones by attacking white-box target malware detectors. The attack adds both \textit{primary} features to bypass malware detection and \textit{side-effect} features to satisfy domain constraints. PiAttack consists of two main phases:
\begin{itemize}
    \item \textbf{Initialization phase.} The first step of PiAttack is to identify the top-n benign features based on the learned weights of the linear SVM. Subsequently, for each benign feature, the attack collects a set of candidate transformations, called gadgets (i.e., slices of the apps' bytecode), by extracting them from benign apps.
    \item \textbf{Attack Phase.} The attack employs a greedy search strategy to find optimal perturbations. It first sorts the collected gadgets based on their feature vector's contribution to the feature vector of the malware app $z$, denoted by $x$. Next, the attack selects the best gadget from the sorted list and combines its feature vector with $x$. This process is repeated until $x$ is classified as benignware. Once the perturbations have been identified, all corresponding gadgets are injected into $z$ to generate RealAE.
\end{itemize} 

Note that for some experiments that need to consider $L_1$ norm bound for PiAttack, we only include specific gadgets that satisfy the bound when combining their feature vector with $x$ in the attack phase. Moreover, in this study, we use DNN instead of linear SVM for AT. Therefore, to clarify the significance of features, we adopt the approach suggested in~\cite{melis2018explaining}, which utilizes the gradients provided by DNN for each feature to explain their global importance. We refer the reader to~\cite{melis2018explaining} for more details about the global explanation of malware detectors. 
It is noteworthy that in this study, PiAttack has been built based on their available source codes published in~\cite{apg_tool}. Considering the recent insights from Pintor et al.~\cite{pintor2022indicators}, PiAttack seems to effectively address several key concerns highlighted in their research.

\begin{itemize}
    \item \textbf{Optimization Challenges and Loss Saturation.} \cite{pintor2022indicators} highlights the issue of loss saturation, where further modifications fail to improve attack success. While PiAttack uses a greedy search rather than gradient-based optimization, it can similarly get stuck in suboptimal solutions. Nevertheless, PiAttack reduces this risk by meticulously choosing gadgets during the initialization phase based on their potential to shift the classification score toward the benign class. This approach ensures the optimization process remains effective, preventing the stagnation that might occur with a less structured search.
    \item \textbf{Feature Contribution and Impact.} PiAttack effectively manages the contribution of each feature. During the initialization phase, gadgets are carefully pre-selected based on their ability to positively influence the classification outcome while avoiding those that might unintentionally reinforce a malicious label. This strategy prevents issues similar to gradient vanishing or exploding, ensuring that each modification significantly contributes to the attack's success without compromising the app's benign appearance.
    \item \textbf{Maintaining Problem-Space Feasibility.} PiAttack is crafted to maintain the original functionality of the Android app. This is accomplished by encapsulating the injected code in conditional statements that are never executed at runtime, ensuring the app's original behavior remains intact. Additionally, PiAttack uses opaque predicates to guard against static analysis tools that might otherwise eliminate the transplanted code. These strategies ensure that the generated adversarial examples remain functional and plausible within the problem space.
\end{itemize}

\section{PK-Feature}
\label{appendix:pk_feature}
This is a white-box attack that iteratively perturbs the impactful features of a malware sample until it reaches the maximum allowable perturbation. It's essential to note that the impactful features are those that exert a more significant influence on the classification outcomes compared to other features, and like PiAttack, they are determined based on the weight parameters of the linear SVM learned during training. Incorporating the dependent features of each perturbed feature, based on our extracted meaningful feature dependencies, enables this attack to generate feature-space RealAEs.

\section{AT with non-uniform perturbations}
\label{apendix:non-uniform-perturbations}
Unlike conventional feature-space adversarial attacks used in AT that can only perturb the training sample under norm-bounded constraints, the main idea of AT with non-uniform perturbations~\cite{b38} is to take into account the data distribution of training data by allowing attackers to generate non-uniform perturbations. Specifically, the paper proposes a new projection approach for the PGD attack as follows:

\begin{equation}
    \begin{split}
   & P(\Omega\delta) = 
      \begin{cases}
      \epsilon\frac{\delta}{\lVert \Omega\delta \rVert_p} & \text{if $\lVert \Omega\delta \rVert_p > \epsilon$}\\
      \delta & otherwise
    \end{cases}\\   
   \end{split}
   \label{eq_projection}
\end{equation}
\noindent where $\lVert . \rVert_p$ shows the $L_p$ norm bound, $\Omega \in \mathbb{R}^{d \times d}$ is a diagonal matrix that can be specified by a \textit{weighted norm}, and $d$ is the dimensions of samples in the training set. By incorporating $\Omega$ in the projection, PGD can perturb important features more than less important features that may have a lesser effect on classification. One of the suggestions of the paper to capture the importance of features is to utilize \textit{Pearson's correlation coefficient} of each feature $f_j$ with the label $y$, which is denoted as $|p_{j,y}|$. Note that $\Omega$ is constructed using this coefficient as follows:

\begin{equation}
     \begin{split}
     \Omega = \frac{diag(\{p_{j,y} ^{-1}\}_{j=1} ^d)}{\lVert diag(\{p_{j,y} ^{-1}\}_{j=1} ^d) \rVert}_2
     \end{split}
     \label{eq:omega}
\end{equation}

\noindent where $p_{j,y} ^{-1} = 1/p_{j,y}$.

\end{document}